\documentclass{article}
\usepackage{ijcai22}

\usepackage{times}
\usepackage{soul}
\usepackage{url}
\usepackage[hidelinks]{hyperref}
\usepackage[utf8]{inputenc}
\usepackage[small]{caption}
\usepackage{graphicx}
\usepackage{amsmath}
\usepackage{amsthm}
\usepackage{booktabs}
\usepackage{algorithm}
\usepackage{algorithmic}
\usepackage{amsfonts}
\usepackage{subfigure}

\usepackage{diagbox}
\usepackage{multirow}
\usepackage{threeparttable}

\title{Personalized Federated Learning with Contextualized Generalization}

\author{
Xueyang Tang$^1$\and
Song Guo$^{1,2}$\footnotemark[1]\And
Jingcai Guo$^{1}$\footnotemark[1]\\
\affiliations
$^1$Department of Computing, The Hong Kong Polytechnic University, Hong Kong, China\\
$^2$The Hong Kong Polytechnic University Shenzhen Research Institute, Shenzhen, China\\
\emails
csxtang@comp.polyu.edu.hk,
song.guo@polyu.edu.hk,
jingcai.guo@gmail.com
}

\begin{document}

\maketitle

\renewcommand{\thefootnote}{\fnsymbol{footnote}}
\footnotetext[1]{Corresponding Authors}

\begin{abstract}
  The prevalent personalized federated learning (PFL) usually pursues a trade-off between personalization and generalization 
  by maintaining a shared global model to guide the training process of local models. 
  However, the sole global model may easily transfer deviated context knowledge 
  to some local models when multiple latent contexts exist across the local datasets.
  In this paper, we propose a novel concept called \textit{contextualized generalization (CG)} to provide each client with fine-grained context knowledge that can better fit the local data distributions and facilitate faster model convergence, based on which we properly design a framework of PFL, dubbed \textit{CGPFL}.
  We conduct detailed theoretical analysis, in which the convergence guarantee is presented and $\mathcal{O}(\sqrt{K})$ speedup over most existing methods is granted. To quantitatively study the generalization-personalization trade-off, we introduce the `generalization error' measure and prove that the proposed \textit{CGPFL} can achieve a better trade-off than existing solutions. Moreover, our theoretical analysis further inspires a heuristic algorithm to find a near-optimal trade-off in \textit{CGPFL}. Experimental results on multiple real-world datasets show that our approach surpasses the state-of-the-art methods on test accuracy by a significant margin.
\end{abstract}

\section{Introduction} 

Recently, personalized federated learning (PFL) has emerged as an alternative to conventional federated learning (FL) to cope with the statistical heterogeneity of local datasets (a.k.a., Non-I.I.D. data). 
%To solve this problem, the personalized federated learning (PFL) has emerged as an alternative to conventional FL to cope with the statistical heterogeneity of datasets (a.k.a., Non-I.I.D. data). 
%
Different from conventional FL that focuses on training a shared global model to explore the global optima of the whole system, i.e., minimizing the averaged loss of clients, the PFL aims at developing a personalized model (distinct from the individually trained local model which usually fail to work due to the insufficient local data and the limited diversity of local dataset) for each client to properly cover diverse data distributions. 
To develop the personalized model, each user needs to incorporate some context information into the local data, since the insufficient local data cannot present the complete context which the personalized model will be applied to~\cite{kairouz19advances}. However, the context is generally latent and can be hardly featurized in practice, especially when the exchange of raw data is forbidden.
In the exsiting PFLs, the latent context knowledge can be considered to be transfered to the local users via the global model update.
During the PFL training, the personalization usually requires personalized models to fit local data distributions as well as possible, while the generalization needs to exploit the common context knowledge among clients by collaborative training. Thus, the PFL is indeed pursuing a trade-off between them to achieve better model accuracy than the traditional FL.
More specifically, the server-side model is trained by aggregating local model updates from each client and hence can obtain the common context knowledge covering diverse data distributions. 
Such knowledge can then be offloaded to each client and contributes to the generalization of personalized models. 

Despite the recent PFL approaches have reported better performance against conventional FL methods, they may still be constrained in personalization by using sole global model as the guidance during the training process. Concretely, our intuition is that: 
% If the feature space is of \textit{significant diversity} across local data distributions, 
If there exists multiple latent contexts across local data distributions, 
then contextualized generalization can provide fine-grained context knowledge and further facilitate the personalized models toward better recognition accuracy and faster model convergence. 
We thus argue one potential bottleneck of current PFL methods is the loss of generalization diversity with only one global model. Worse still, the global model may also easily degrade the overall performance of PFL models due to negative knowledge transfers between the disjoint contexts.

In this paper, we design a novel PFL training framework, dubbed \textit{CGPFL}, by involving the proposed concept, i.e., \textit{contextualized generalization (CG)}, to handle the challenge of the context-level heterogeneity. 
More specifically, we suppose the participating clients can be covered by several latent contexts based on their statistical characteristics and each latent context can be corresponded to a generalized model maintained in the server. 
The personalized models are dynamically associated with the most pertinent generalized model and guided by it with fine-grained contextualized generalization in an iterative manner. 
We formulate the process as a bi-level optimization problem considering both the global models with contextualized generalization maintained in the server and the personalized models trained locally in clients. 

The main contributions of this work are summarize as follows:
\begin{itemize}

\item 
To the best of our knowledge, we are the first to propose the concept of \textit{contextualized generalization (CG)} to provide fine-grained generalization and seek a better trade-off between personalization and generalization in PFL, and further formulate the training as a bi-level optimization problem that can be solved effectively by our designed \textit{CGPFL} algorithm.

\item 
We conduct detailed theoretical analysis to provide the convergence guarantee and prove that \textit{CGPFL} can obtain a $\mathcal{O}(\sqrt{K})$ times acceleration over the convergence rate of most existing algorithms for non-convex and smooth case. We further derive the generalization bound of \textit{CGPFL} and demonstrate that the proposed \textit{contextualized generalization} can constantly help reach a better trade-off between personaliztion and generalization in terms of generalization error against the state-of-the-arts. 
%for the personalized models, so as to achieve a better performance. We formulate the PFL with clustered generalization as a bi-level optimization problem and design an effective algorithm called CGPFL to solve it.
%\item 
%In theoretical analysis, we provide detailed convergence analysis of \textit{CGPFL}. Especially, \textit{CGPFL} obtains a $\mathcal{O}(\sqrt{K})$ acceleration over the convergence rate of most existing FL algorithms for non-convex and smooth case. What's more, we derive the generalization bound of \textit{CGPFL} and compare it with that of the state-of-the-art methods, which shows that clustered generalization can help reach a better trade-off between personaliztion and generalization in terms of generalization error.

\item 
We provide a heuristic improvement of \textit{CGPFL}, dubbed \textit{CGPFL-Heur}, by minimizing the generalization bound in the theoretical analysis, to find a near-optimal trade-off between personalization and generalization. \textit{CGPFL-Heur} can achieve a near-optimal accuracy with negligible additional computation in the server, while retaining the same convergence rate as that of \textit{CGPFL}.

\item
Experimental results on multiple real-world datasets demonstrate that our proposed methods, i.e., \textit{CGPFL} and \textit{CGPFL-Heur}, can achieve higher model accuracy than the state-of-the-art PFL methods in both convex and non-convex cases.

\end{itemize}

\section{Related Work}

% \subsection{Clustered Federated learning}
Considering that one shared global model can hardly fit the heterogeneous data distributions, some recent FL works~\cite{ghosh20CFL,sattler20bi-CFL,briggs20FL+HC,mansour20three} try to cluster the participating clients into multiple groups and develop corresponding number of shared global models by aggregating the local updates. After the training process, the obtained global models are offloaded to the corresponding clients for inference.
Since these methods only reduce the FL training into several sub-groups, of which each global model is still shared by their in-group clients, the personalization is scarce and the offloaded models can still hardly cover the heterogeneous data distributions across the in-group clients. 
%FL framework 
Specifically, \textit{IFCA}~\cite{ghosh20CFL} requires each client to calculate the losses on all global models to estimate its cluster identity during each iteration, and result in significantly higher computation cost. 
%In \textit{IFCA} \citep{ghosh20CFL}, each client has to calculate the loss values on all global models to estimate its cluster identity at each iteration. Therefore, the computation cost on clients is significantly higher compared with other FL methods. 
\textit{CFL}~\cite{sattler20bi-CFL} demonstrates that the conventional FL even cannot converge in some Non-I.I.D. settings and provides intriguing perspective for clustered FL with bi-partitioning clustering. However, it can only work for some special Non-I.I.D. case described as \textit{`same feature \& different labels'}~\cite{hsieh20noniid}. 
%and lack the generalization ability to some other application scenarios. 
%%
\textit{FL+HC}~\cite{briggs20FL+HC} divides the clients clustering and the model training processes separately, and only conducts the clustering once at a manually defined step, while the training remains the same as conventional FL. 
%and conducts the clustering once at a certain step (which needs to be defined manually), while the training process remains the same as conventional FL. 
% Differently, robust FL against the Byzantine machine in Non-I.I.D. case is studied in~\cite{ghosh19robust}, where the {\itshape k}-Means algorithm is utilized to cluster the clients and then find out the outlier (Byzantine) machines. 
%
Last, three effective PFL approaches are proposed in~\cite{mansour20three}, of which the user clustering method is very similar to \textit{IFCA}~\cite{ghosh20CFL}.

% \subsection{Personalized Federated Learning}https://www.overleaf.com/project/61418f1dc6ffbed704aaa636
Most recently, the PFL approaches have attracted increasing attention~\cite{kairouz19advances}. 
%emerged as one of the most promising approaches for the statistical challenge of Non-I.I.D. data in federated learning and has attracted increasing attention \citep{QiangY21PFL, kairouz19advances, li20flsurvey, kulkarni2020survey}. 
%From the perspective of context featurization, 
%some context features that indicate the personal characteristics of the clients are required for the training of personalized models \citep{hard18, wang19percontext}. However, due to the requirement for additional personal features, this category of works can hardly be applied to more general cases. 
Among them, a branch of works~\cite{hanzely20mix,hanzely20lower,deng20peradap} propose to mix the global model on the server with local models to acquire the personalized models. 
Specifically, Hanzely \textit{et al.}~\cite{hanzely20lower,hanzely20mix} formulate the mixture problem as a combined optimization of the local and global models, while \textit{APFL}~\cite{deng20peradap} straightforwardly mixes them with an adaptive weight. \textit{FedMD}~\cite{li2019fedmd} exploits the knowledge distillation (KD) to transfer the generalization information to local models and allows the training of heterogeneous models in FL setting. 
Differently, \textit{FedPer}~\cite{arivazhagan19fedper} splits the personalized models into two separate parts, of which the base layers are shared by all the clients and trained on the server, and the personalization layers are trained to adapt to individual data and maintain the privacy properties on local devices. 
\textit{MOCHA}~\cite{smith17fmtl} considers the model training on the clients as relevant tasks and formulate this problem as a distributed multi-task learning objective. 
%which can address the statistical and systematic challenges in FL setting. However, \textit{MOCHA} only considers the convex objective with strict duality.
% that is rarely applicable in real-world application. 
% Jiang \textit{et al.}~\cite{jiang19permeta} and 
Fallah \textit{et al.}~\cite{fallah20perFed} make use of the model agnostic meta learning (\textit{MAML})
% ~\cite{finn17maml} 
to implement the PFL, of which the obtained meta-model contains the generalization information and can be utilized as a good initialization point of training.
 % (only one or few steps to convergence) 
%of personalized models.

\section{Problem Formulation}

We start by formalizing the FL task and then introduce our proposed method. Given $N$ clients and the their Non-I.I.D. datasets $\widetilde{D}_1, ..., \widetilde{D}_i, ..., \widetilde{D}_N$ that subject to the underlying distributions as $D_1,...,D_i,...,D_N$ ($D_i \in \mathbb{R}^{d \times n_i}$ and $i \in [N]$). 
%on the clients are denoted as $\tilde{D}_1, ..., \tilde{D}_i, ..., \tilde{D}_N$ which are subject to the underlying distributions $D_1,...,D_i,...,D_N$ ($D_i \in R^{d \times n_i}$ and $i \in [N]$), respectively. 
Every client $i$ has $m_i$ instances $z^{i,j} = (\mathbf{x^{i,j}}, y^{i,j})$, $j \in [m_i]$, where $\mathbf{x}$ is the data features and $y$ denotes the label. Hence, the objective function of the conventional FL can be described as~\cite{li2021ditto}:
\begin{small}
\begin{equation}
\min_{\omega \in R^d}\{ G(\omega) := G\big(f_1(\omega; \widetilde{D}_1),...,f_N(\omega; \widetilde{D}_N)\big) \},
\end{equation}
\end{small}where $\omega$ is the global model and $f_i: \mathbb{R}^d \rightarrow \mathbb{R}, i \in [N]$ denotes the expected loss function over the data distribution of client $i$: $f_i(\omega;\widetilde{D}_i)=\mathbb{E}_{z^{i,j} \in \widetilde{D}_i}[\tilde{f}_i(\omega;z^{i,j})]$.
% \begin{small}
% \begin{equation}
% f_i(\omega; \tilde{D}_i) = \mathbb{E}_{z^{i,j} \in \tilde{D}_i}[\widetilde{f}_i(\omega; z^{i,j})].
% \end{equation}
% \end{small}
$G(\cdot)$ denotes the aggregation method to obtain the global model $\omega$. For example, \textit{FedAvg}~\cite{mcmahan17} applies $G(\omega) = \sum_{i=1}^N \frac{m_i}{m}f_i(\omega)$ to do the aggregation, where $m$ is the total number of instances on local devices. %i.e., $m = \sum_{i=1}^N m_i$. 
%$G(\omega)$ can be expressed as $G(\omega) = \sum_{i=1}^N \frac{m_i}{m}f_i(\omega)$ in \textit{FedAvg} \citep{mcmahan17}, where $m$ is the total number of samples on all the local devices, i.e., $m = \sum_{i=1}^N m_i$.
% while $G(\omega) = \sum_{i=1}^N {\lambda}_i f_i(\omega)$ in AFL (\cite{mohri19AFL}), where $\sum_{i=1}^N {\lambda}_i = 1$ and ${\lambda}_i\geq 0$ for all $i \in [N]$.
% In most PFL frameworks \cite{hanzely20mix, hanzely20lower, dinh20pFedMe, fallah20perFed, li2021ditto, deng20peradap}, the objective functions can be decomposed into two parts, of which one is solved on the server by aggregating the uploaded model updates from the local devices to generate the global model, and the other is settled parallelly on the clients to get the personalized models. The training process of each personalized model is guided via a regularizer in the local objective.

To handle the challenge of rich statistical diversities in PFL, especially in the cases where the local datasets belong to several latent contexts, our \textit{CGPFL} propose to maintain $K$ context-level generalized models in the server to guide the training of personalized models on the clients. 
%(Figure~\ref{fig:framework}). 
%
During training, the local training process based on its local dataset can \textit{push} the personalized model to fit its local data distribution as well as possible. Meanwhile, the regularizer will dynamically \textit{pull} the personalized model as close as possible to the most pertinent generalized model during the iterative algorithm, from which the fine-grained context knowledge can be transferred to each personalized model to better balance the generalization and personalization. 
%
%On the one hand, the local training based on the local dataset pushing the personalized model to fit its local data distribution as well as possible. On the other hand, the regularizer will pulling the personalized model as close as possible to the nearest generalized model which is dynamically chosen during the iterative algorithm. 
%Thanks to the fine-grained common knowledge transferred from the nearest generalized model, the obtained personalized models can balance the generalization and personalization much better. 
% which furture improves the model accuracy on test dataset.
% Considering that the Moreau envelop has two useful properties: \textit{smoothness} and \textit{fast proximal minimization}, which can facilitate several learning algorithm designs \cite{dinh20pFedMe, zhou19moreau}, we adopt this form of regularized loss function with $L_2$-norm as the local objective. It is noticed that a similar form is also utilized in Hanzely \textit{et al.} \cite{hanzely20lower, hanzely20mix}. Thus, 
Hence, the overall objective function of \textit{CGPFL} can be described as a bi-level optimization problem as:
\begin{small}
\begin{equation*}
    \label{equ:client1}
    % \min_{[{\theta}_1,...,{\theta}_N] \in R^{d\times N}} \frac{1}{N} \sum_{i=1}^{N} \{ F_i({\theta}_i) := f_i({\theta}_i)+{\lambda}r({\theta}_i,{\omega}^{\ast}_k) \}, wherei \in C_k
    \min_{{\Theta} \in \mathbb{R}^{d\times N}} \frac{1}{N} \sum_{i=1}^{N} \Big\{ F_i({\theta}_i) := f_i({\theta}_i)+{\lambda}r({\theta}_i,{\omega}^{\ast}_k) \Big\}, i \in C_k^{\ast},
\end{equation*}
\begin{equation*}
  \label{equ:gen}
  s.t.\quad {\Omega}^{\ast}, C_K^{\ast} = \mathop{\arg\min}\limits_{{\Omega} \in \mathbb{R}^{d \times K}, C_K} G({\omega}_1,...,{\omega}_K;C_K),
\end{equation*}
\end{small}where $\theta_i$ ($i \in [N]$) denotes the personalized model on client $i$ and $\Theta = [\theta_1,...,\theta_N]$. The context-level generalized models are denoted by $\Omega = [\omega_1,...,\omega_K]$. $\lambda$ is a hyper-parameter and $C_k$ denotes the corresponding context that client $i$ belongs to. Considering the latent contexts are represented in disjoint subspaces respectively, the function $G(\cdot)$ can be decomposed as $G({\omega}_1,...,{\omega}_K;C_K) = \frac{1}{K}\sum_{k=1}^K G_k(\omega_k;C_k)$.
%indicates which cluster the client $i$ belongs to (i.e., which generalized model the personalized model $\theta_i$ should be associated with).  

In general, there exists two alternative strategies to generate the context-level generalized models. 
The intuitive one is to solve the inner-level objective $\min_{\Omega \in \mathbb{R}^{d\times K}}G({\omega}_1,...,{\omega}_K)$ based on local datasets, which is similar to \textit{IFCA}~\cite{ghosh20CFL}. 
%In this way, the existing \textit{CFL} methods, for example \textit{IFCA} in \citep{ghosh20CFL} can be adopted to solve the inner-level objective. 
However, the computation overhead is high in the local devices while their available computation resources are usually limited. 
%%on which the available computation resources are usually limited, especially when they are edge devices.
% Worse still, uploading the original local gradients is also accompanied with higher risk of privacy leakage~\cite{lyu20FLthreats,zhu20deepleakage}. 
%
Comparing the local objective that trains a generalized model $\omega_k$ based on local dataset, i.e., ${\omega}_i^{\ast} = \mathop{\arg\min}\limits_{\omega}f_i(\omega;\widetilde{D}_i)$, with that of the personalized model, i.e., $\theta_i^{\ast} = \mathop{\arg\min}\limits_{\theta_i}\{f_i(\theta_i;\widetilde{D}_i)+\lambda r(\theta_i,\omega_k^{\ast})\}$, we notice that the locally obtained $\theta_i^{\ast}$ can be regarded as the distributed estimation of $\omega_k^{\ast}$. 
In this way, the regularizer $r(\theta_i^{\ast},\omega_k^{\ast})$ can be used to evaluate the estimation error, and we can further derive the context-level generalized models by minimizing the average estimation error. 
In this paper, we use $L2$-norm i.e., $r({\theta}_i,{\omega}_k) = \frac{1}{2}{\lVert{\theta}_i-{\omega}_k\rVert}^2$ as the regularizer, which is also adopted in various prevalent PFL methods~\cite{hanzely20mix,hanzely20lower,dinh20pFedMe,li2021ditto} and has empirically demonstrated to be superior over other regularizers, e.g., the symmetrized KL divergence in~\cite{li2021ditto}. 
Hence, we formulate our overall objective as:
%of our work as following:
\begin{small}
\begin{equation*}
    \label{equ:client2}
    % \min_{[{\theta}_1,...,{\theta}_N] \in R^{d\times N}} \frac{1}{N} \sum_{i=1}^{N} \{ F_i({\theta}_i) := f_i({\theta}_i)+{\lambda}r({\theta}_i,{\omega}^{\ast}_k) \}, wherei \in C_k
    \min_{{\Theta} \in \mathbb{R}^{d\times N}} \frac{1}{N} \sum_{i=1}^{N} \Big\{ F_i({\theta}_i) := f_i({\theta}_i)+\frac{\lambda}{2}{\lVert{\theta}_i-{\omega}^{\ast}_k\rVert}^2 \Big\}, i \in C_k^{\ast},
\end{equation*}
\begin{equation}
\label{equ:km}
s.t.\quad {\Omega}^{\ast}, C_K^{\ast} = \mathop{\arg\min}\limits_{{\Omega} \in \mathbb{R}^{d \times K}, C_K} \sum_{k=1}^{K} q_k\sum_{j \in C_k} p_{k,j}{\lVert {\theta}_j - {\omega}_k\rVert}^2,
\end{equation}
\end{small}

We adopt $p_{k,j}=\frac{1}{\lvert C_k \rvert}$ and  $q_k=\frac{\lvert C_k \rvert}{N}$ in this paper, where $C_k (k \in [K])$ denotes the latent and disjoint context $k$, and $\lvert C_k \rvert$ is the number of clients that belong to the context $k$. 
% Evidently, the objective~\eqref{equ:km} is the special case of the objective~\eqref{equ:gen}, in which $G(\omega_1,...,\omega_K;C_K)=$
Intriguingly, the inner-level objective is exactly the classic objective of 
% {\itshape k}-Means clustering~\cite{lloyd1982kmeans,arthur2006kmeans++}. 
$k$-means clustering~\cite{lloyd1982kmeans}. 
We notice that when $K = 1$, the above objective is equivalent to the overall objective in~\cite{dinh20pFedMe}, which means that the objective in~\cite{dinh20pFedMe} can be regarded as a \textit{special case} ($K = 1$) of ours.

\section{Design of \textit{CGPFL}}

% \subsection{Overview}
In this section, we introduce our proposed \textit{CGPFL} in detail. The key idea is to dynamically relate the clients to $K$ latent and disjoint contexts based on their uploaded local model updates, and then develop a generalized model for each context by aggregating the updates in each uaer group. These generalized models are utilized to guide the training directions of personalized models and transfer contextualized generalization to them. 
Both the personalized models and the generalized models are trained in parallel, so we can denote the model parameters in matrix form. The generalized models can be written as ${\Omega}_K := [\omega_1, \dots, \omega_k, \dots, \omega_K] \in \mathbb{R}^{d\times K}$, and the corresponding local approximations are ${\Omega}_{I,R} := [\tilde{\omega}_{1,R}, \dots, \tilde{\omega}_{i,R}, \dots, \tilde{\omega}_{N,R}]$,
% $ \in \mathbb{R}^{d\times N}$
where $R$ is the number of local iterations and $\tilde{\omega}_{i,R}, \omega_k \in \mathbb{R}^d, \forall i \in [N], k \in [K]$.
%% Putting this definition in the appendix
% For simplicity, we define $G_k(\omega_k) = \sum_{j \in C_k} \frac{1}{\lvert C_k \rvert} {\lVert {\theta}_i - {\omega}_k\rVert}^2$ and $G_K(\Omega_K) = [G_1(\omega_1),...G_i(\omega_i),...,G_K(\omega_K)]$. Similarly, $G_i(\omega_k) = 2r(\theta_i,\omega_k) = {\lVert {\theta}_i - {\omega}_k\rVert}^2$.
In this paper, we use capital characters to represent matrices unless stated otherwise.

\subsection{\textit{CGPFL}: Algorithm}

We design an effective alternating optimization framework to minimize the overall objective in~\eqref{equ:km}. Specifically, the upper-level problem can be decomposed into $N$ separate sub-problems with fixed generalized models and to be solved on local devices in parallel. 
Next, we can further settle the inner-level problem to derive the generalized models with fixed personalized models. 
Since the solution to the sub-problems of the upper-level objective has been well-explored in recent PFL methods~\cite{dinh20pFedMe,li2021ditto,hanzely20lower}, we hereby mainly focus on the inner-level problem. 
%in the remaining part of this section. 
%explaining the designed approach to solving the inner-level problem (correspoding to forming the generalized models) in the remaining part of this section.
% As regard to solving the inner-level objective in~\eqref{equ:km},
We alternately update the context-level generalized models $\Omega_K$ and the context indicator $C_K$ to obtain the optimal generalized models. We view the personalized models, i.e., $\Theta_I=[\theta_i,...,\theta_N]$, as private data, and distributionally update the context-level generalized models $\Omega_K$ on clients with fixed context indicator $C_K$. 
During each server round, the server conducts $k$-means clustering on uploaded local parameters ${\Omega}_{I,R}^t$ to cluster each client into $K$ disjoint contexts, and the clustering results $C_K$ are re-arranged to the matrix form as $P^t \in \mathbb{R}^{N\times K}$.
For example, if client $i, i\in [N]$ is clustered into the context $C_j, j\in [K]$ (where $C_j, j\in [K]$ are sets, the union $\mathop{\bigcup}_{j\in [K]}C_j$ and intersection $\mathop{\bigcap}_{j\in [K]}C_j$ are the set $[N]$ and empty set, respectively), the element $(P^t)_{i, j}$ is defined as $\frac{1}{\lvert C_j \rvert}$, or set $0$ otherwise. 
%Otherwise, the elements are set $0$. 
In this way, the elements of every column in $P^t$ amount to $1$, i.e. $\sum_{i=1}^N (P^t)_{i,j} = 1 \text{, } \forall j, t.$ 
% \begin{small}
% \begin{equation}
% \label{sum_Pt}
%   \sum_{i=1}^N (P^t)_{i,j} = 1 \text{, } \forall j, t.
% \end{equation}
% \end{small}
%\vspace{-0.37cm}

\begin{algorithm}[htb]
\begin{footnotesize}
\caption{\textit{CGPFL}: Personalized Federated Learning with Contextualized Generalization}
\label{algo:pFedKM}
\textbf{Input}: ${\Theta}_I^0, {\Omega}_K^0, P^0, T, R, S, K, \lambda, \eta, \alpha, \beta$. \\
% \textbf{Parameter}: Optional list of parameters\\
\textbf{Output}: ${\Theta}_I^T$.
\end{footnotesize}
\begin{footnotesize}
\begin{algorithmic}[1]
  \FOR {$t=0$ to $T-1$} %\qquad \qquad \qquad \qquad \qquad \qquad \qquad \qquad \qquad \qquad \qquad Global communication rounds
    % \STATE Server sends ${\Omega}_K^t$ to clients according to ${\Omega}_I^t={\Omega}_K^t J_K^t$, where $J_K^t P^t = I_K$.
    \STATE Server sends ${\Omega}_K^t$ to clients according to $P^t$.
    \FOR {local device $i=1$ to $N$ in parallel} %\qquad \qquad \qquad \qquad \qquad \qquad \qquad \qquad \qquad \qquad \qquad Parallel
      % \STATE Initialization: ${\Omega}_{I,0}^t = {\Omega}_I^t$.
      \STATE Initialization: ${\Omega}_{I,0}^t = {\Omega}_K^tJ^t$.
      \STATE Local update for the sub-problem of $G({\Theta}_I, {\Omega}_K)$: 
      \FOR {$r=0$ to $R-1$} %\qquad \qquad \qquad \qquad \qquad \qquad \qquad \qquad \qquad \qquad Local update rounds
        % \STATE Local solver find an approximate solution $\tilde{\theta}_i({\omega}_{i,r}^t)$:
        \FOR {$s=0$ to $S-1$}
            \STATE Update personalized model: ${\theta}_i^{s+1} = {\theta}_i^s - \eta \nabla F_i({\theta}_i^s)$.
            % \STATE ${\theta}_i^{s+1} = {\theta}_i^s - \eta \nabla F_i({\theta}_i^s)$.
        \ENDFOR
        % \STATE Local update: ${\omega}_{i,r+1}^t = {\omega}_{i,r}^t-\beta\nabla G({\omega}_{i,r}^t)$.
        \STATE Local update: $\tilde{\omega}_{i,r+1}^t=\tilde{\omega}_{i,r}^t - \beta\nabla_{\omega_i}G(\tilde{\theta}_i(\tilde{\omega}_{i,r}^t), \tilde{\omega}_{i,r}^t)$.
        % \STATE ${\omega}_{i,r+1}^t = {\omega}_{i,r}^t-\beta \nabla G\Big(r\big(\tilde{\theta}_i({\omega}_{i,r}^t),{\omega}_{i,r}^t\big)\Big)$.
        % \STATE ${\omega}_{i,r+1}^t = {\omega}_{i,r}^t-\beta\nabla G({\omega}_{i,r}^t)$.
      \ENDFOR
    \ENDFOR
    \STATE Clients send back $\tilde{\omega}_{i,R}^t$ and server conducts clustering (e.g., $k$-means++) on models ${\Omega}_{I,R}^t$ to obtain $P^{t+1}$.
    % \STATE Server conducts (\textit{k}-means++) clustering on models ${\Omega}_{I,R}^t$ to obtain $P^{t+1}$.
    \STATE Global aggregation: ${\Omega}_K^{t+1} = {\Omega}_K^t - \alpha({\Omega}_K^t - {\Omega}_{I,R}^t P^{t+1})$.
    % \STATE ${\Omega}_K^{t+1} = {\Omega}_K^t - \alpha({\Omega}_K^t - {\Omega}_{I,R}^t P^{t+1})$.
  \ENDFOR
  \STATE {\textbf{return} The personalized models ${\Theta}_I^T$.}
\end{algorithmic}
\end{footnotesize}
\end{algorithm}
% \vspace{-0.37cm}

When considering the relationship between the consecutive $P^t$, we can formulate the iterate as $P^{t+1} = P^t Q^t$, where $Q^t\in \mathbb{R}^{K\times K}$ is a square matrix. We can find that to maintain 
% the property ~\eqref{sum_Pt} 
the above property
of $P^t$ ($\forall t$), the matrix $Q^t$ must satisfies that: 
% $\sum_{j=1}^K (Q^t)_{j,k} = 1 \text{, } \forall k, t$ and  $\sum_{k=1}^K (Q^t)_{j,k} = 1 \text{, } \forall j, t.$
\begin{footnotesize}
\begin{equation}
\sum_{j=1}^K (Q^t)_{j,k} = 1 \text{, } \forall k, t \quad\quad \text{ and } \quad\quad \sum_{k=1}^K (Q^t)_{j,k} = 1 \text{, } \forall j, t.
\end{equation}
\end{footnotesize}It is noticed that the clustering is based on the latest model parameters ${\Omega}_{I}^{t+1}$ that depends on ${\Omega}_{I}^{t}$, and the latest gradient updates given by clients. Hence, $P^{t+1}$ is determined by and only by $P^t$ and $Q^t$. Then we can consider this global iteration as a discrete-time Markov chain and $Q^t$ corresponds the transition probability matrix. 

During each local round, the clients 
need to first utilize local datasets to solve the regularized optimization objective, i.e., the upper-level objective in~\eqref{equ:km} with fixed $\tilde{\omega}_{i,r}^t$ to obtain a $\delta$-approximate solution $\tilde{\theta}_i(\tilde{\omega}_{i,r}^t)$. 
% The optimization objective ~\eqref{equ:client} is the well-known Moreau envelope function and can be solved efficiently by many existing approaches. Here, we utilize the method exploited in \cite{dinh20pFedMe} to obtain the $\delta$-approximate solution.
Then, each client is required to calculate the gradients $\nabla_{\omega_i}G(\tilde{\theta}_i(\tilde{\omega}_{i,r}^t), \tilde{\omega}_{i,r}^t)$ with fixed $\tilde{\theta}_i(\tilde{\omega}_{i,r}^t)$ and update the model using 
$\tilde{\omega}_{i,r+1}^t=\tilde{\omega}_{i,r}^t - \beta\nabla_{\omega_i}G(\tilde{\theta}_i(\tilde{\omega}_{i,r}^t), \tilde{\omega}_{i,r}^t)$
, where $\beta$ is the learning rate and 
% the gradient $\nabla_{\omega_i}G(\tilde{\theta}_i(\omega_{i,r}^t), \omega_{i,r}^t)$ can be calculated by 
$\nabla_{\omega_i}G(\tilde{\theta}_i(\tilde{\omega}_{i,r}^t), \tilde{\omega}_{i,r}^t) = \frac{2}{N}\nabla r\big(\tilde{\theta}_i(\tilde{\omega}_{i,r}^t),\tilde{\omega}_{i,r}^t \big)$.
To reduce the communication overhead, our \textit{CGPFL} allows the clients to process several iterations before uploading the latest model parameters to the server. 
The details of \textit{CGPFL} is given in algorithm~\ref{algo:pFedKM}, from which we can summarize the parameters update process as:
\begin{small}
\begin{equation}
  \label{iterates}
  {\Omega}_{I,R}^{t-1} \mathop{\longrightarrow}^{P^t} {\Omega}_K^t
  \mathop{\longrightarrow}^{J^t} {\Omega}_{I,0}^t 
  \mathop{\longrightarrow}^{H_I^t} {\Omega}_{I,R}^t 
  \mathop{\longrightarrow}^{P^{t+1}} {\Omega}_K^{t+1},
\end{equation}
\end{small}where $P^{t+1} = P^t Q^t$ and $J^t P^t = I_K$ ($J^t\in \mathbb{R}^{K\times N}$ and $I_K$ is an identity matrix), $\forall t$.

\subsection{Convergence Analysis}
%In this section, we provide the convergence analysis of the proposed \textit{CGPFL}. 
Since the inner-level objective in~\eqref{equ:km} is non-convex, we focus on analyzing the convergence rate under the smooth case. 
% Based on the parameters update process given in~\eqref{iterates}, 
Firstly, we can write the local updates as: 
\begin{small}
\begin{equation}
  \label{local_updates}
  {\Omega}_{I,R}^t = {\Omega}_{I,0}^t-\beta R H_I^t,
\end{equation}
\end{small}where $H_I^t = \frac{1}{R}\sum_{r=0}^{R-1} H_{I,r}^t$ and $H_{I,r}^t = \frac{2}{N}\big({\Omega}_{I,r}^t - \widetilde{\Theta}_I({\Omega}_{I,r}^t) \big)$.
 % can be considered as the biased estimation of the gradient $\nabla G_I({\Omega}_{I,r}^t)$.
% , because $\mathbb{E}\big[ H_{I,r}^t \big] \neq \nabla G_I({\Omega}_{I,r}^t)$. 
Based on~\eqref{local_updates} and the update process in~\eqref{iterates}, we can obtain the global updates as:
\begin{small}
\begin{flalign*}
  {\Omega}_K^{t+1} &= (1-\alpha){\Omega}_K^t + \alpha{\Omega}_{I,R}^t P^{t+1} &
  \\&= {\Omega}_K^t[(1 - \alpha)I_K + \alpha Q^t] - {\alpha\beta R} H_I^t P^t Q^t. &
%\end{displaymath}
\end{flalign*}
\end{small}\textbf{Definition 1} ($L$-smooth) {\itshape (i.e., L-Lipschitz gradient)} {\itshape If a function $f$ satisfies  $\lVert\nabla f(\omega) - \nabla f({\omega}^\prime)\rVert \leq L\lVert\omega - (\omega)^\prime\rVert$, $\forall \omega$, ${\omega}^\prime$, we say $f$ is $L$-smooth.}

\noindent\textbf{Assumption 1} (smoothness) 
% {\itshape The loss functions $f_i, \forall i$ and $r(\cdot)$ are $L_f$-smooth and $L_r$-smooth, respectively.}
{\itshape The loss functions $f_i$ is $L$-smooth and $G(\omega_k)$ is $L_G$-smooth, $\forall i$, $k$.}

\noindent\textbf{Assumption 2} (bounded intra-context diversity) 
{\itshape The variance of local gradients to the corresponding context-level generalized models is upper bounded by:}
\begin{small}
\begin{equation}
    \frac{1}{\lvert C_k \rvert}\sum_{i \in C_k}{\lVert \nabla G_{k,i}(\omega_k) - \nabla G_k(\omega_k) \rVert}^2 \leq \delta_G^2, \forall k \in [K],
\end{equation}
\end{small}where $G_{k,i}(\omega_k) := r({\theta}_i, {\omega}_k)$.

\noindent\textbf{Assumption 3} (bounded parameters and gradients) 
{\itshape The generalized model parameters ${\Omega}_K^t$ 
and the gradients $\nabla G_K({\Omega}_K^t)$ 
are upper bounded by ${\rho}_{\Omega}$ 
and ${\rho}_g$, respectively.}
\begin{small}
\begin{gather}
  {\big\lVert {\Omega}_K^t \big\rVert}^2 \leq {\rho}_{\Omega}^2 \qquad 
  \text{and} 
  \qquad {\big\lVert \nabla G_K({\Omega}_K^t) \big\rVert}^2 \leq {\rho}_g^2 \text{, } \quad \forall t
\end{gather}
\end{small}where ${\rho}_{\Omega}$ and ${\rho}_{g}$ are finite non-negative constants, and $\nabla G_K({\Omega}_K^t) := [\nabla G_1({\omega}_1^t),...,\nabla G_k({\omega}_k^t),...,\nabla G_K({\omega}_K^t)]$.
% \begin{equation}
%   {\big\lVert {\Omega}_K^t \big\rVert}^2 \leq {\rho}_{\Omega}^2,
% \end{equation}

\noindent\textbf{Proposition 1} {\itshape~\cite{dinh20pFedMe}} 
{\itshape The deviation between the $\delta$-approximate and the optimal solution is upper bounded by $\delta$. That is:}
\begin{small}
\begin{equation}
  \mathbb{E}\Big[ {\big\lVert \widetilde{\Theta}_I({\Omega}_{I,r}^{t}) - \widehat{\Theta}_I({\Omega}_{I, r}^{t}) \big\rVert}^2\Big] \leq N\delta^2, \forall r, t,
\end{equation}
\end{small}where $\widetilde{\Theta}_I$ is the $\delta$-approximate solution and $\widehat{\Theta}_I$ is the matching optimal solution.

Assumption 1 provides typical conditions for convergence analysis, and assumption 2 is common in analyzing algorithms that are built on SGD. As for assumption 3, the model parameters are easily bounded by using projection during the model training process, while the gradients can be bounded with the smooth condition and bounded model parameters. 
% Based on these assumptions, we first give a transitional result on the convergence condition. 
To evaluate the convergence of the proposed \textit{CGPFL}, we adopt the technique used in~\cite{dinh20pFedMe} to define that: 
\begin{small}
\begin{equation*}
  \mathbb{E}\Big[\frac{1}{K}{\big\lVert \nabla G_K({\Omega}_K^{t^\ast})\big\rVert}^2\Big] := \frac{1}{T}\sum_{t=0}^{T-1}\mathbb{E}\Big[\frac{1}{K}{\big\lVert \nabla G_K({\Omega}_K^t)\big\rVert}^2\Big],
\end{equation*}
\end{small}where $t^{\ast}$ is uniformly sampled from the set $\{0, 1, \dots, T-1 \}$. 

%\noindent\textbf{Lamma 1} (Convergence condition of {\itshape pFedKM})

\noindent\textbf{Theorem 4.1} (Convergence of \textit{CGPFL}) 
Suppose Assumption 1, 2 and 3 hold. If $\beta \leq \frac{1}{2\sqrt{R(R+1)L_G^2}}, \forall R\geq 1$, $\alpha \leq 1$, and ${\hat{\alpha}}_0 := \min\Big\{ \frac{8{\alpha}^2{\rho}_{\Omega}^2}{K{\Delta}_G}, \sqrt{\frac{4}{3}}\frac{\alpha\rho_{\Omega}}{\rho_{g}}, \sqrt{\frac{1}{416{L_G}^2}}\alpha \Big\}$, where ${\Delta}_G$ is defined as ${\Delta}_G := \mathbb{E}\Big[\frac{1}{K}\sum_{k=1}^{K}G_k({\omega}_k^{0}) - \frac{1}{K}\sum_{k=1}^{K}G_k({\omega}_k^{T}) \Big]$, we have:
\begin{itemize}
\item The convergence of the generalized models:
% \begin{small}
% \begin{equation*}
% \begin{aligned}
% &\mathbb{E}\Big[\frac{1}{K}{\big\lVert \nabla G_K({\Omega}_K^{t^\ast})\big\rVert}^2\Big] \leq \mathcal{O}\bigg( \mathbb{E}\Big[\frac{1}{K}{\big\lVert \nabla G_K({\Omega}_K^{t^\ast})\big\rVert}^2\Big] \bigg) := \\
% &\mathcal{O}\bigg( \frac{ 12{\alpha}^2 ({\rho}_{\Omega}^2/K)}{{\hat{\alpha}_0}^2 T} + \frac{80(6 ({\rho}_{\Omega}^2/K) L_G^2 \delta^2 )^{\frac{1}{2}}}{\sqrt{NKRT}} + \frac{ 64N{\delta}^2}{K} \bigg).
% \end{aligned}
% \end{equation*}
% \end{small}
\begin{small}
\begin{flalign*}
&\frac{1}{K}\mathbb{E}\Big[{\big\lVert \nabla G_K({\Omega}_K^{t^\ast})\big\rVert}^2\Big] &
\\&\leq \mathcal{O}\bigg( \frac{ 48{\alpha}^2 ({\rho}_{\Omega}^2/K)}{{\hat{\alpha}_0}^2 T} + \frac{80(26 ({\rho}_{\Omega}^2/K) L_G^2 \delta^2 )^{\frac{1}{2}}}{\sqrt{NKRT}} + \frac{ 52{\delta}^2}{KN} \bigg). &
\end{flalign*}
\end{small}
\item The convergence of the personalized models:
\begin{small}
\begin{flalign*}
&\frac{1}{N}\sum_{i=1}^{N}\mathbb{E}\Big[ {\big\lVert \widetilde{\Theta}_I^{t^{\ast}} - \Omega_K^{t^{\ast}}J^{t^{\ast}} \big\rVert}^2\Big] &
\\&\leq \mathcal{O}\Big( \frac{1}{K}\mathbb{E}\Big[{\big\lVert \nabla G_K({\Omega}_K^{t^\ast})\big\rVert}^2\Big] \Big) + \mathcal{O}\Big( \frac{\delta_G^2}{\lambda^2} + \delta^2 \Big).
\end{flalign*}
\end{small}
\end{itemize}
%\begin{equation*}
%  \mathbb{E}\Big[\frac{1}{K}{\big\lVert \nabla F_K({\Omega}_K^{t^\ast})\big\rVert}^2\Big] \leq \mathcal{O}\bigg( \mathbb{E}\Big[\frac{1}{K}{\big\lVert \nabla F_K({\Omega}_K^{t^\ast})\big\rVert}^2\Big] \bigg) :=
%\end{equation*}
%\begin{equation}
%  \mathcal{O}\bigg( \frac{ K{\beta}^2 {\rho}_{\Omega}^2}{2{\alpha_0}^2 T} + \frac{(2^R N K)^{\frac{1}{2}}\lambda\delta L_F{\rho}_{\Omega}}{R\sqrt{T}} + 16N{\lambda}^2{\delta}^2 \bigg)
%\end{equation}
\noindent \textbf{Remark 4.1} Theorem 1 shows that the proposed \textit{CGPFL} can achieve a convergence rate of $\mathcal{O}\big( 1/\sqrt{KNRT} \big)$, which is $\mathcal{O}(\sqrt{K})$ times faster than most of the state-of-the-art works~\cite{karimireddy20SCAFFOLD,deng20peradap,reddi2020adaptive} that achieved (i.e., $\mathcal{O}\big( 1/\sqrt{NRT} \big)$) in non-convex FL setting. 
The detailed proof of convergence rate is provided in the \underline{\textbf{\textit{Appendix}}} of this paper.

\subsection{Generalization Error}
We analyse the generalization error of \textit{CGPFL} in this section. Before starting the analysis, we first introduce two important definitions as follows.

\noindent \textbf{Definition 2} (Complexity) {\itshape Let $\mathcal{H}$ be a hypothesis class (correspanding to ${\omega} \in \mathbb{R}^d$ in neural network), and $\lvert D \rvert$ be the size of dataset $D$, the complexity of $\mathcal{H}$ can be expressed by the maximum disagreement
between two hypotheses on a dataset $D$:}
\begin{equation}
{\lambda}_{\mathcal{H}}(D) = \sup_{h_1, h_2 \in \mathcal{H}}\frac{1}{\lvert D \rvert}\sum_{(x,y)\in D} \lvert h_1(x)-h_2(x) \rvert.
\end{equation}
\noindent \textbf{Definition 3} (Label-discrepancy) {\itshape Consider a hypothesis class $\mathcal{H}$, the label-discrepancy between two data distributions $D_1$ and $D_2$ is given by:}
\begin{equation}
disc_{\mathcal{H}}(D_1, D_2) = \sup_{h \in \mathcal{H}}\lvert \mathcal{L}_{D_1}(h)-\mathcal{L}_{D_2}(h) \rvert,
\end{equation}
where $\mathcal{L}_{D}(h)=\mathbb{E}_{(x,y) \in D}[\mathit{l}(h(x), y)]$.

\noindent\textbf{Theorem 4.2} (Generalization error of \textit{CGPFL}) When Assumption 1 is satisfied, with probability at least $1-\delta$, the following holds:
\begin{small}
\begin{flalign*}
&\sum_{i=1}^N \frac{m_i}{m}\Big\{ \mathcal{L}_{D_i}(\hat{h}_i^{\ast}) - \min_{h \in \mathcal{H}}\mathcal{L}_{D_i}(h) \Big\} &
\\&\leq 2\sqrt{\frac{\log{\frac{N}{\delta}}}{m}} + \sqrt{\frac{dK}{m}\log{\frac{em}{d}}} + (\lambda + \frac{L}{2})cost({\Theta}^{\ast}, {\Omega}^{\ast}; K) &
\\&+ \sum_{i=1}^{N}\frac{m_i}{m}\big\{ 2B\mathop{\lambda_{\mathcal{H}}}(D_i) + \mathop{disc(D_i,\widetilde{D}_i)} \big\}, &
\end{flalign*}
\end{small}where $B$ is a positive constant with $\big\lvert\mathcal{L}_{D}(h_1) - \mathcal{L}_{D}(h_2)\big\rvert \leq B\mathop{\lambda_{\mathcal{H}}}(D)$, $\forall h_1 \text{, } h_2 \in \mathcal{H}$.
% The dataset $D_{C_k}$ is the union of $D_j$, $j \in C_k$, i.e., $D_{C_k} = \mathop{\bigcup}_{j \in C_k}D_j$. 
Besides, $\hat{h}_i^{\ast}$ is given by $\hat{h}_i^{\ast} = \mathop{\arg\min}\limits_{\theta_i}\big\{ \mathcal{L}_{\widetilde{D}_i}(h(\theta_i)) + {\lVert \theta_i - \omega_k^{\ast} \rVert}^2 \big\}$ and $cost({\Theta}^{\ast}, {\Omega}^{\ast}; K) = \sum_{i=1}^{N}\frac{m_i}{m}\min_{k \in[K]}{\lVert {\theta}_i^{\ast} - {\omega}_k^{\ast} \rVert}^2$.

\noindent \textbf{Remark 4.2} Theorem 2 gives the generalization error bound of \textit{CGPFL}. When $K = 1$, it yields the error bound of PFL with single global model~\cite{li2021ditto,dinh20pFedMe,hanzely20mix,hanzely20lower}. As the number of contexts increases, the second terms become larger, while the last term get smaller. Hence, our \textit{CGPFL} can alwalys reach better personalization-generalization trade-off by adjusting the number of contexts $K$, and further achieve higher accuracy than the existing PFL methods. The detailed proof of generalization error is given in the \underline{\textbf{\textit{Appendix}}} of this paper. 
% (please refer to the \textit{Supplementary Material} folder).

% \subsection{\textit{CGPFL-Heur}: Heuristic Improvement of \textit{CGPFL}}
\subsection{\textit{CGPFL-Heur}: The Heuristic Improvement}
As discussed, Theorem 2 indicates that there exists a optimal $K^{\ast}$ ($K^{\ast} \in [K]$) to achieve the minimal generalization error that corresponds to the highest model accuracy. 
Theoretically, the optimal $K^{\ast}$ can be obtained by minimizing the generalization bound in Theorem 2. We can find that the first and the third term have no relationship with the number of latent contexts, that is, they are irrelevant to $K$. 
Therefore, we can obtain an optimal $K^{\ast}$ by minimizing the following expression:
\begin{small}
\begin{equation}
\label{objective:K}
e(K) := \sqrt{\frac{dK}{m}\log{\frac{em}{d}}} + \mu\cdot cost({\Theta}^{\ast}, {\Omega}^{\ast}; K),
\end{equation}
\end{small}where $\mu$ is a hyper-parameter which is induced by the unknown constant $L$. 
The above objective can be solved in the server along with the clustering.
% and will not bring about extra computation overhead on clients. 
%Because clustering is conducted in the server, the above objective is also solved in the server. 
In the down-to-earth experiments, we notice that the latent context structure can be learned efficiently in the first few rounds. Based on this observation, we believe that \textit{CGPFL-Heur} can efficiently figure out a near-optimal solution $\hat{K}$ by operating the solver of~\eqref{objective:K} only in the first few rounds (in the experimental part, we only operate the solver in the first global round), and after that, the obtained $\hat{K}$ will no longer be updated. 
%In the related experiments, we make \textit{CGPFL-Heur} operate the solver only in the first global round, and after that, the obtained $\hat{K}$ will no longer be updated. 
In this way, \textit{CGPFL-Heur} can reach a near-optimal trade-off (corresponding to the near-optimal $\hat{K}$) between generalization and personalization with negligible additional computation in the server. 
Moreover, in view of the fact that we only need to operate the solver in the first few rounds, \textit{CGPFL-Heur} can retain the same convergence rate as \textit{CGPFL}.

\section{Experiments}
%In this section, we introduce the experiments and show the performance of our proposed \textit{CGPFL} and \textit{CGPFL-Heur} on several real-world datasets including MNIST \citep{lecun1998gradient}, CIFAR10 \citep{krizhevsky2009CIFAR10} and Fashion-MNIST (FMNIST) \citep{xiao2017FMNIST}, compared with the FL algorithm \textit{FedAvg}, one typical CFL algorithm and several state-of-the-art PFL algorithms. 
%Besides, we also provide some experimental results that support the convergence analysis and then claim a reasonable overhead caused by the {\itshape k}-Means clustering at the server. More details of the experimental setup are provided in the \textit{Supplementary Material}.

\subsection{Experimental Setup}

% \subsubsection{Datasets}
\textbf{Dataset Setup:} 
Three datasets including MNIST~\cite{lecun1998gradient}, CIFAR10~\cite{krizhevsky2009CIFAR10}, and Fashion-MNIST (FMNIST)~\cite{xiao2017FMNIST} are used in our experiments. To generate Non-I.I.D. datasets for each client, we split the whole dataset as follows. 1) MNIST: we distribute the train-set containing $60,000$ digital instances into $40$ clients, and each of them is only provided with 3 classes out of total 10. The number of instances obtained by each client is randomly chosen from the range of $[400, 5000]$, of which $75\%$ are used for training and the remaining $25\%$ for testing. 2) CIFAR10: We distribute the whole dataset containing $60,000$ instances into $40$ clients, and each of them is also provided with 3 classes out of total 10. The number of instances obtained by each client is randomly chosen from the range of $[400, 5000]$. The train/test remains $75\%$/$25\%$. 3) Fashion-MNIST: a more challenging replacement of MNIST, the Non-I.I.D. splitting is the same as MNIST. 

\textbf{Competitors:} 
We compare our \textit{CGPFL} and \textit{CGPFL-Heur} with seven state-of-the-art works: one traditional FL method, \textit{FedAvg}~\cite{mcmahan17}; one typical cluster-based FL method, \textit{IFCA}~\cite{ghosh20CFL}; and five most recent PFL models, \textit{APFL}~\cite{deng20peradap}, \textit{Per-FedAvg}~\cite{fallah20perFed}, \textit{L2SGD}~\cite{hanzely20mix}, \textit{pFedMe}~\cite{dinh20pFedMe}, and \textit{Ditto}~\cite{li2021ditto}.
%with the traditional FL methods, \textit{FedAvg} \citep{mcmahan17}; one typical CFL method, \textit{IFCA} \citep{ghosh20CFL}; and several state-of-the-art PFL methods, including \textit{APFL} \citep{deng20peradap}, \textit{Per-FedAvg} \citep{fallah20perFed}, \textit{L2SGD} \citep{hanzely20mix}, \textit{pFedMe} \citep{dinh20pFedMe}, and \textit{Ditto} \citep{li2021ditto}.

% \subsubsection{Learning models}
\textbf{Model Architectures:} 
%Although our convergence analysis is devised for the non-convex case because of the limitation that the classic {\itshape k}-Means objective is non-convex. In this paper, we also conduct experiments for convex loss function. 
1) For strongly convex case, we use a $l_2$-regularized multinomial logistic regression model (MLR) with the softmax and cross-entropy loss, in line with~\cite{dinh20pFedMe};
2) For the non-convex case, we apply a neural network with one hidden layer of size $128$ and a softmax layer at the end (DNN) for evaluation. In addition, we apply a CNN that has two convolutional layers and two fully connected layers for the CIFAR10. 
All competitors and our \textit{CGPFL} and \textit{CGPFL-Heur} are based on the same configuration and fine-tuned to their best performance. 
%We compare our \textit{CGPFL} and \textit{CGPFL-Heur} with the state-of-the-art FL, CFL and most importantly PFL algorithms in different cases, 
%based on the same and fine-tuned hyper-parameters (e.g., learning rates, batch sizes, and number of local and global iterations).

\begin{table}[htbp]
\centering
\scalebox{0.77}{
\begin{threeparttable}
\caption{Comparison of test accuracy. We set $N=40$, $\alpha=1$, $\lambda=12$, $S=5$, $lr=0.005$ and $T=200$ for MNIST and Fashion-MNIST (FMNIST), and $T=300$, $lr=0.03$ for CIFAR10, where $lr$ denotes the learning rate.}  
\label{tab:acc_overall}
\setlength{\tabcolsep}{2.7mm}{    
\begin{tabular}{lcccccccc}  
\toprule  
\multirow{2}{*}{Method}
&\multicolumn{2}{c}{MNIST}
&\multicolumn{2}{c}{FMNIST}
&\multicolumn{2}{c}{CIFAR10}\cr   
\cmidrule(lr){2-3} 
\cmidrule(lr){4-5} 
\cmidrule(lr){6-7}
  &MLR  &DNN   &MLR  &DNN  &CNN    \cr              
\midrule
        \textit{FedAvg} & $88.63$ & $91.05$ & $82.44$ & $83.45$ & $46.34$ \cr
        %\hline
        \textit{IFCA} ($K=4$) & $95.27$ & $96.19$ & $91.55$ & $92.56$ &$60.22$ \cr
        %\hline
        \textit{L2SGD} & $89.46$ & $92.48$ & $88.59$ & $90.64$ & $58.68$ \cr
        %\hline
        \textit{APFL} & $92.69$ & $95.59$ & $92.60$ & $93.76$ & $72.12$ \cr
        %\hline
        \textit{pFedMe (PM)} & $91.90$ & $92.20$ & $85.49$ & $86.87$ & $68.88$ \cr
        %\hline
        \textit{Per-FedAvg (HF)} & $92.44$ & $93.54$ & $87.17$ & $87.57$ & $71.46$ \cr
        %\hline
        \textit{Ditto} & $89.96$ & $92.85$ & $88.62$ & $90.56$ & $69.56$ \cr
        %\hline
        \textbf{\textit{CGPFL} ($\mathbf{K=4}$)} & \underline{$95.65$} & \underline{$96.55$} & \underline{$92.65$} & \underline{$93.56$} & \underline{$72.78$} \cr
        %\hline
        \textbf{\textit{CGPFL-Heur}} & $\mathbf{97.41}$ & $\mathbf{98.03}$ & $\mathbf{95.18}$ & $\mathbf{96.00}$ & $\mathbf{74.75}$ \cr
\bottomrule  
\end{tabular}
}
%\footnotesize{The.}
\end{threeparttable}
}
%\footnotesize{The}
\vspace{-0.27cm}
\end{table}

% \subsection{Overall Performance of \textit{CGPFL} and \textit{CGPFL-Heur}}
\subsection{Overall Performance}

\begin{figure}[htbp]
  \centering
  \subfigure[acc-MNIST-MLR]{
    \begin{minipage}{0.22\textwidth}
    \centering
    \label{acc-MNIST-MLR}
    \includegraphics[width=1.1\textwidth]{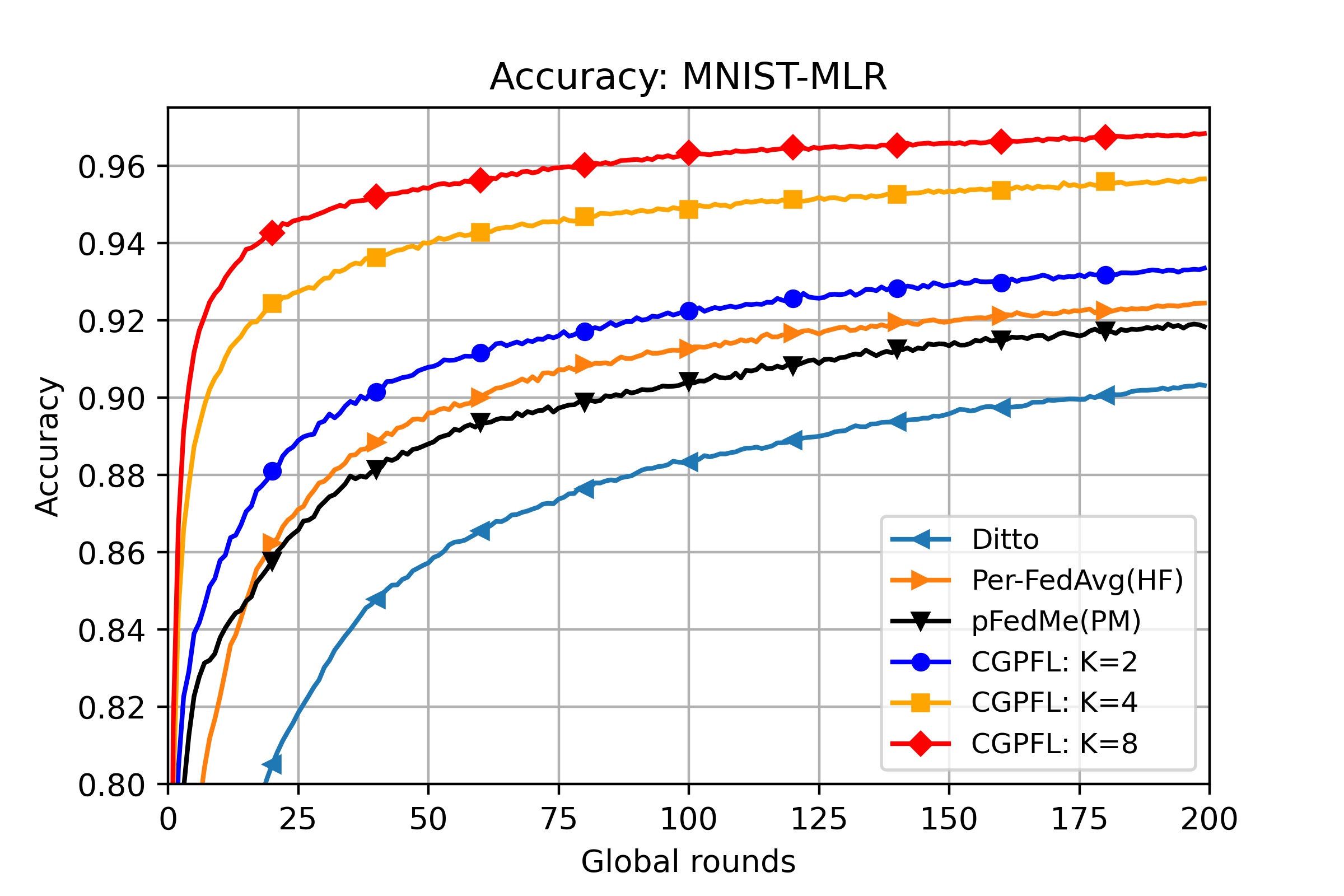}
    \end{minipage}
  }
  \subfigure[acc-MNIST-DNN]{
    \begin{minipage}{0.22\textwidth}
    \centering
    \label{acc-MNIST-DNN}
    \includegraphics[width=1.1\textwidth]{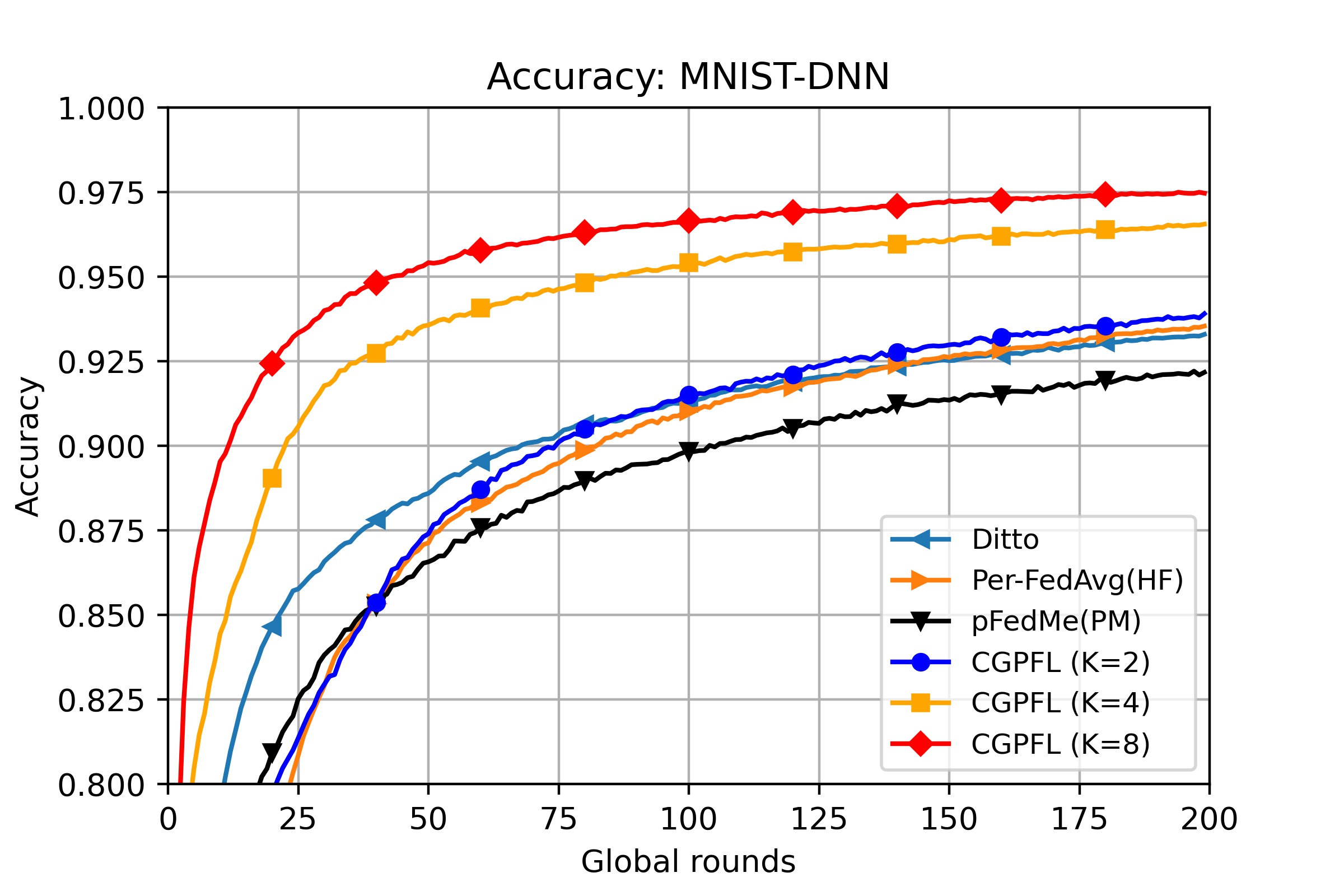}
    \end{minipage}
  }

  \subfigure[loss-MNIST-MLR]{
    \begin{minipage}{0.22\textwidth}
    \centering
    \label{loss-MNIST-MLR}
    \includegraphics[width=1.1\textwidth]{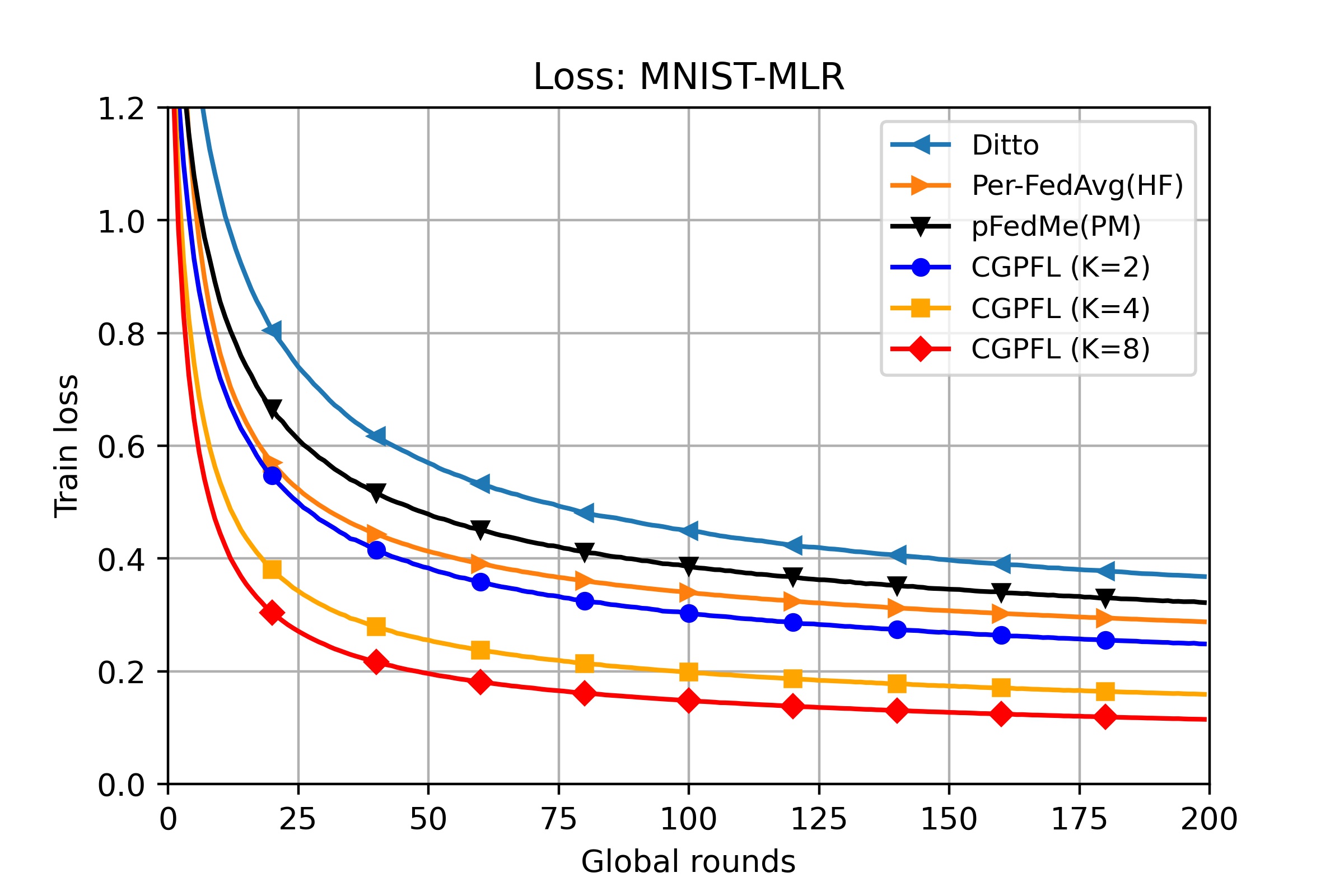}
    \end{minipage}
  }
  \subfigure[loss-MNIST-DNN]{
    \begin{minipage}{0.22\textwidth}
    \centering
    \label{loss-MNIST-DNN}
    \includegraphics[width=1.1\textwidth]{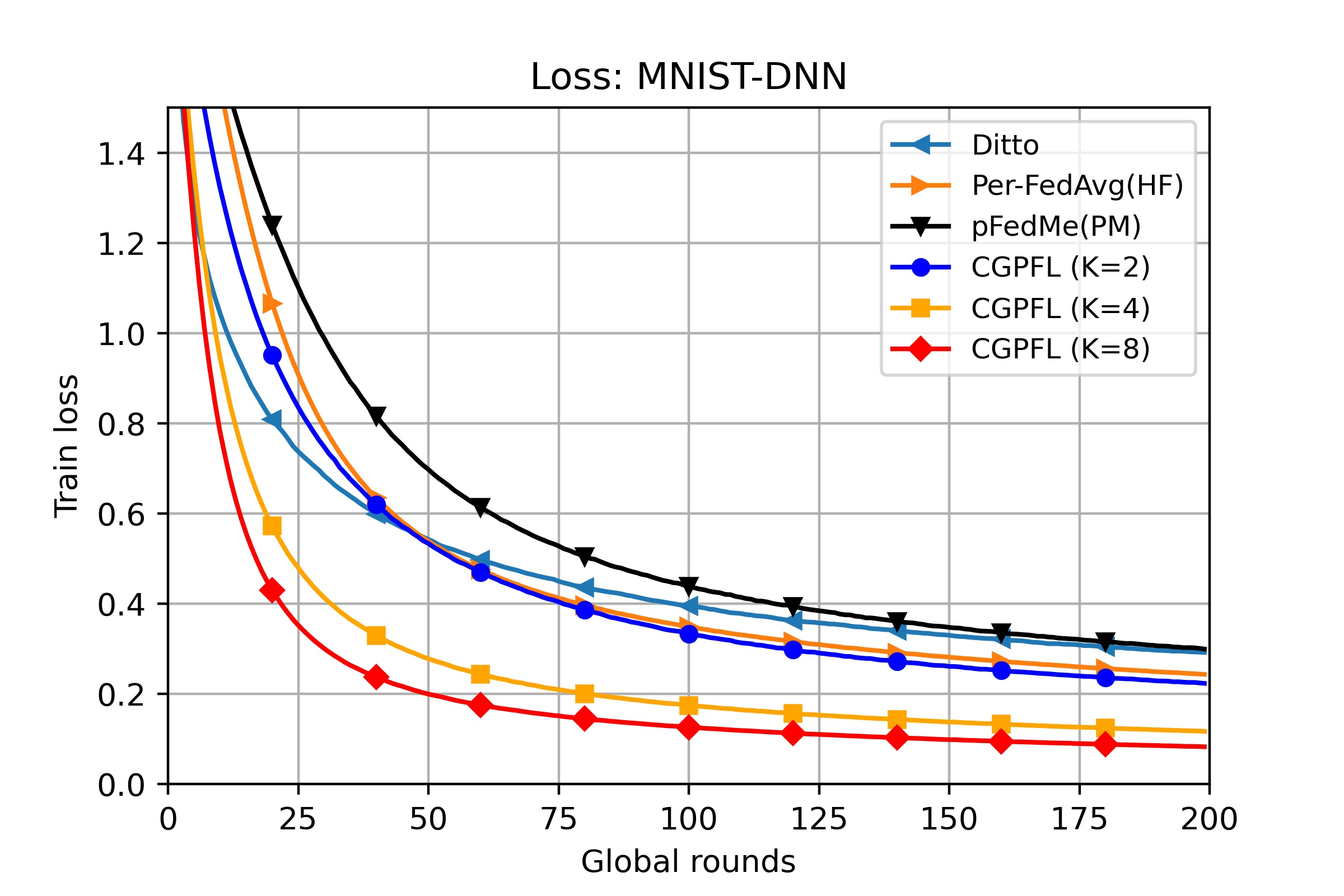}
    \end{minipage}
  }
  \caption{Performance on MNIST for different $K$ with $N=40$, $\alpha=1$, $\lambda=12$, $R=10$, $S=5$.}
  \label{fig:mnist-K}
  %\Description{}
  \vspace{-0.27cm}
\end{figure}

\begin{figure}[htbp]
  \centering
  \subfigure[acc-FMNIST-MLR]{
    \begin{minipage}{0.22\textwidth}
    \centering
    \label{acc-FMNIST-MLR}
    \includegraphics[width=1.1\textwidth]{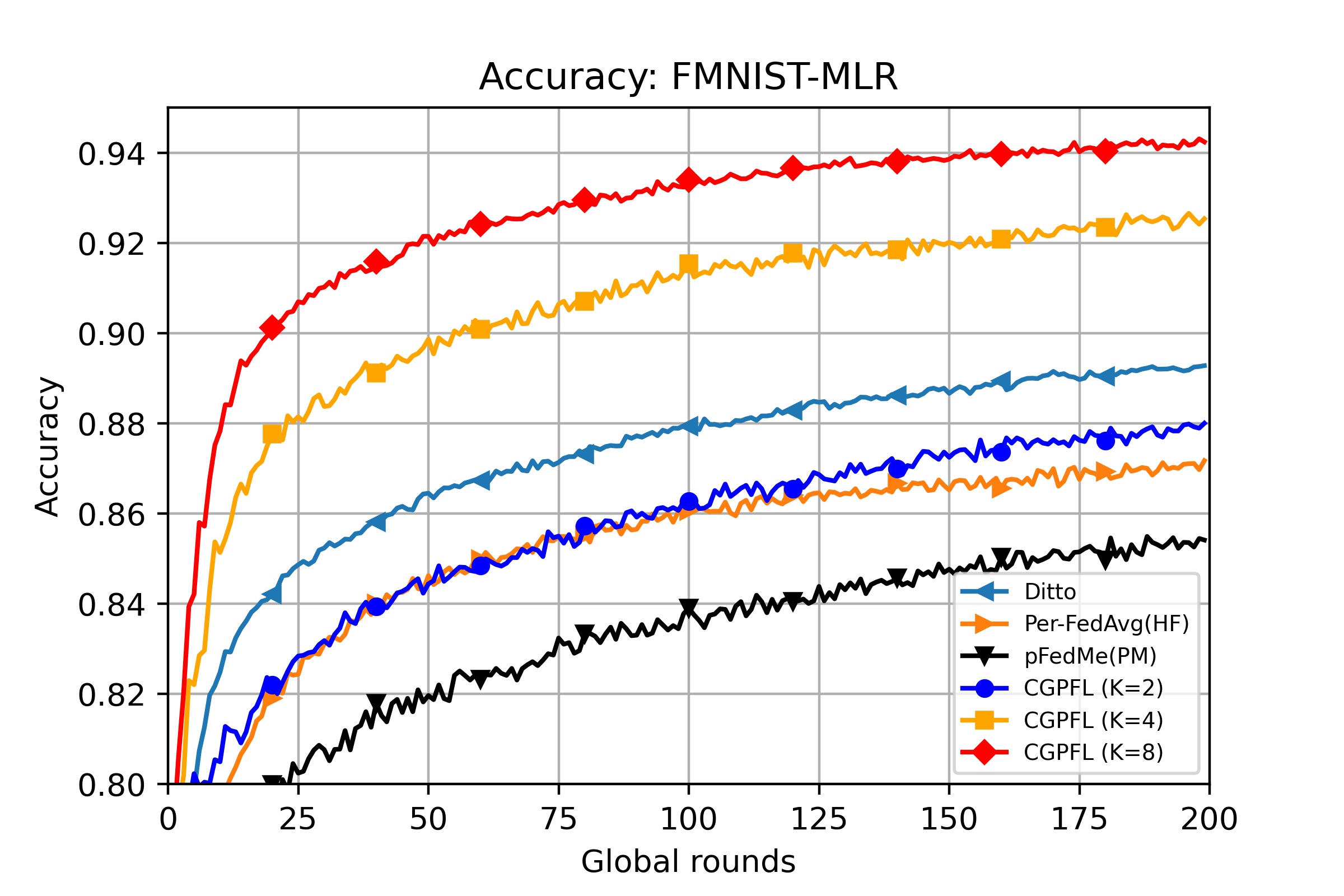}
    \end{minipage}
  }
  \subfigure[acc-FMNIST-DNN]{
    \begin{minipage}{0.22\textwidth}
    \centering
    \label{acc-FMNIST-DNN}
    \includegraphics[width=1.1\textwidth]{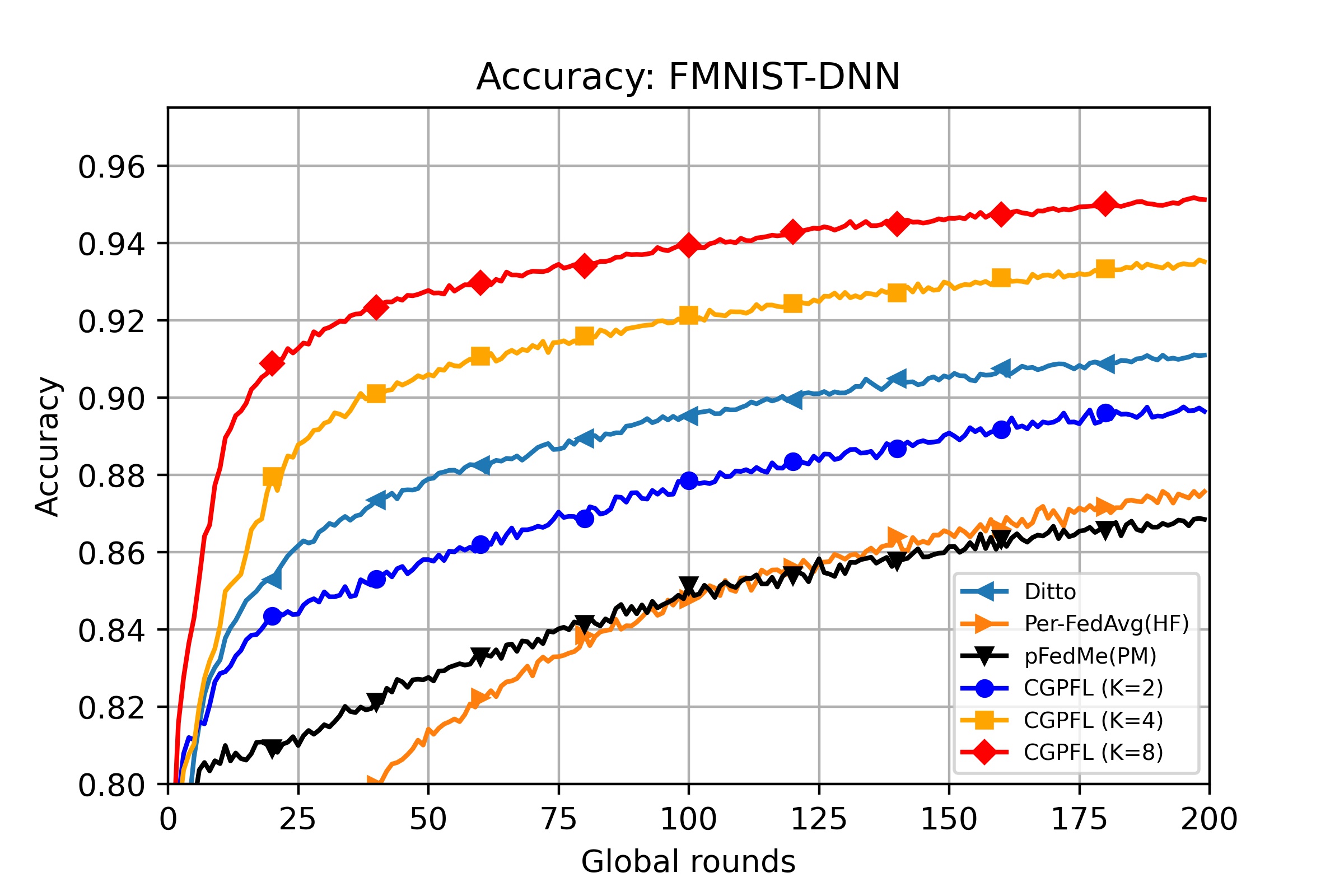}
    \end{minipage}
  }

  \subfigure[loss-FMNIST-MLR]{
    \begin{minipage}{0.22\textwidth}
    \centering
    \label{loss-FMNIST-MLR}
    \includegraphics[width=1.1\textwidth]{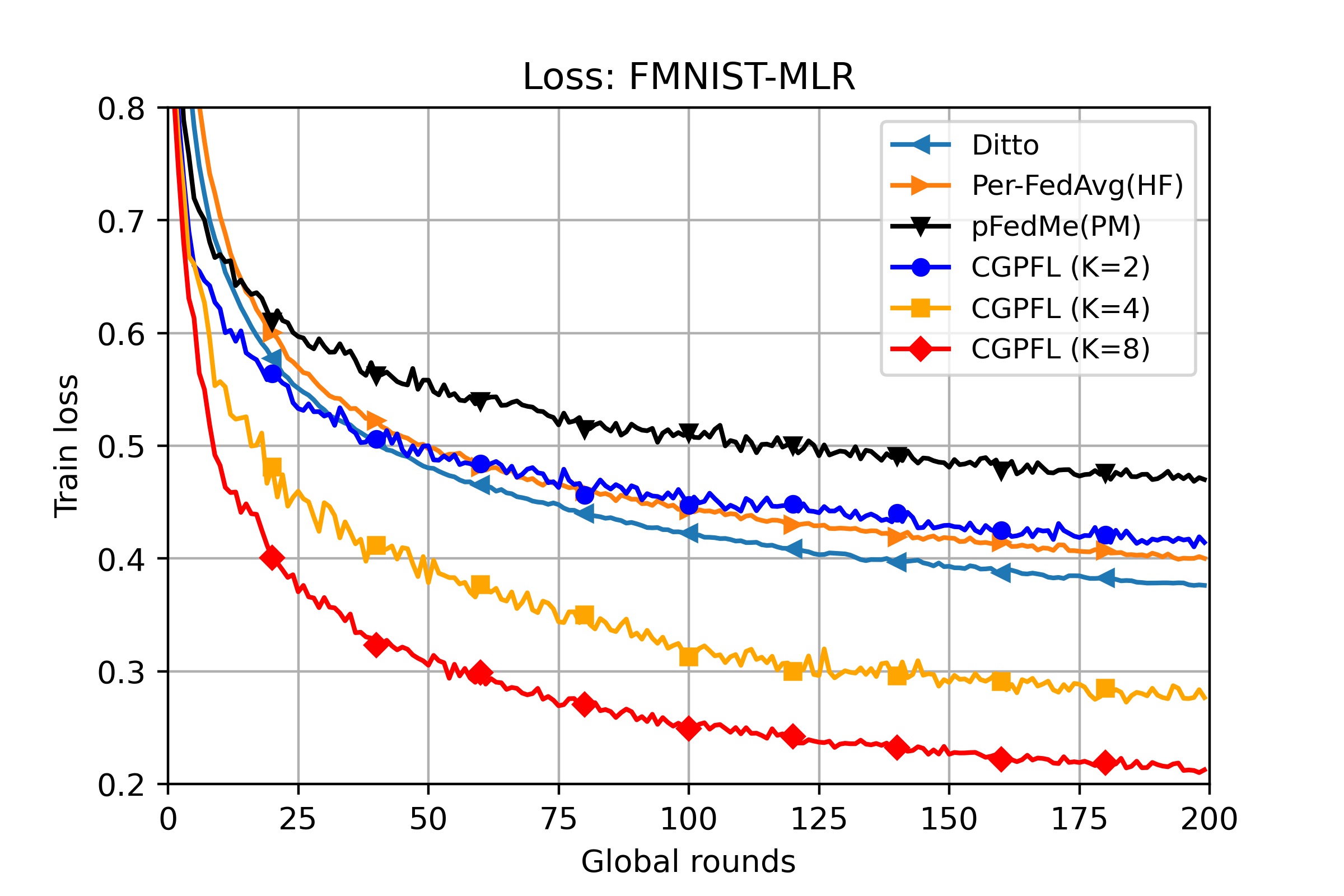}
    \end{minipage}
  }
  \subfigure[loss-FMNIST-DNN]{
    \begin{minipage}{0.22\textwidth}
    \centering
    \label{loss-FMNIST-DNN}
    \includegraphics[width=1.1\textwidth]{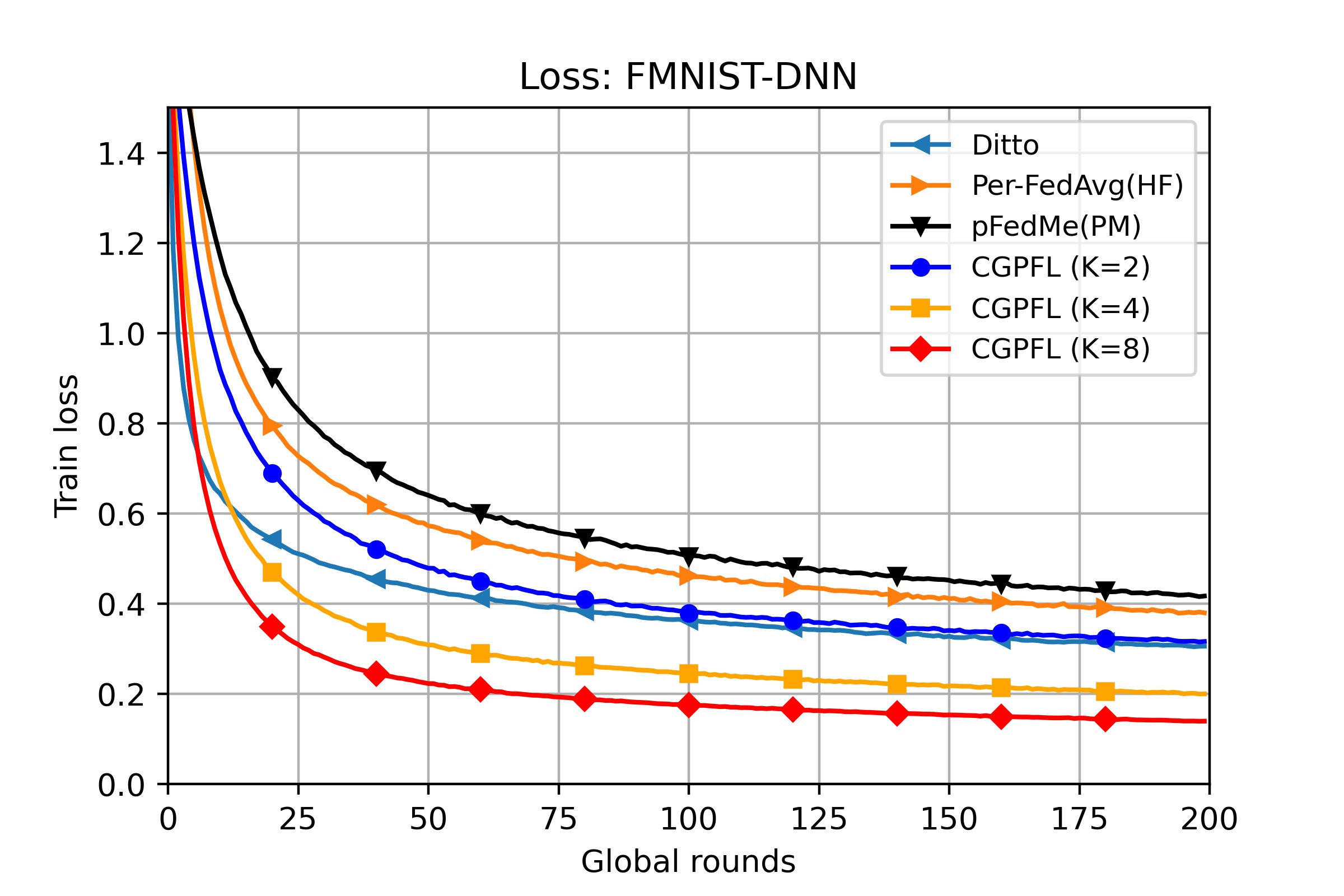}
    \end{minipage}
  }
  \caption{Performance on FMNIST for different $K$ with $N=40$, $\alpha=1$, $\lambda=12$, $R=10$, $S=5$.}
  \label{fig:fmnist-K}
  %\Description{}
  \vspace{-0.27cm}
\end{figure}

The comprehensive comparison results of our \textit{CGPFL} and \textit{CGPFL-Heur} are shown in Table~\ref{tab:acc_overall}. 
%on several real-world datasets are demonstrated in Table~\ref{tab:acc_overall}. 
It can be observed that our methods outperform the competitors with large margins for both non-convex and convex cases on all datasets, even if \textit{IFCA} works with a good initialization.
%In this section, we will compare the performance of our algorithm 
%{\itshape pFedKM} with that of the state-of-the-art algorithm 
%{\itshape pFedMe} \cite{dinh20pFedMe} and {\itshape IFCA} \cite{ghosh20CFL} on several real-world datasets and the synthetic dataset, for nonconvex and convex cases. Firstly, the comprehensive results are given in Table ~\ref{tab:acc}. We can see that our algorithm {\itshape pFedKM} outperforms {\itshape pFedMe} and {\itshape IFCA} for both nonconvex and convex cases on diverse datasets even if {\itshape IFCA} works with good initialization. 
Besides, although we only provide the proof of convergence rate under non-convex case, as shown in Figure~\ref{fig:mnist-K} and Figure~\ref{fig:fmnist-K}, the extensive experiments further demonstrate that our methods constantly obtain better performance against multiple state-of-the-art PFL metohds (\textit{pFedMe}, \textit{Ditto}, and \textit{Per-FedAvg}) with faster convergence rate under both strongly-convex and non-convex cases. 
%
%Of course, too large $K$ should not be adopted to avoid overfitting.
Specifically, the figures in Figure~\ref{fig:mnist-K} show the results for MNIST dataset on MLR and DNN model, while the figures in Figure~\ref{fig:fmnist-K} give the results for Fashion-MNIST dataset on MLR and DNN model.

\subsection{Further evaluation on \textit{CGPFL-Heur}}
\begin{figure}[t]
\vspace{-0.27cm}
  \centering
  \subfigure[\textit{CGPFL} with variable $K$]{
    \begin{minipage}{0.22\textwidth}
    \centering
    \label{acc-optimal-K}
    \includegraphics[width=1.1\textwidth]{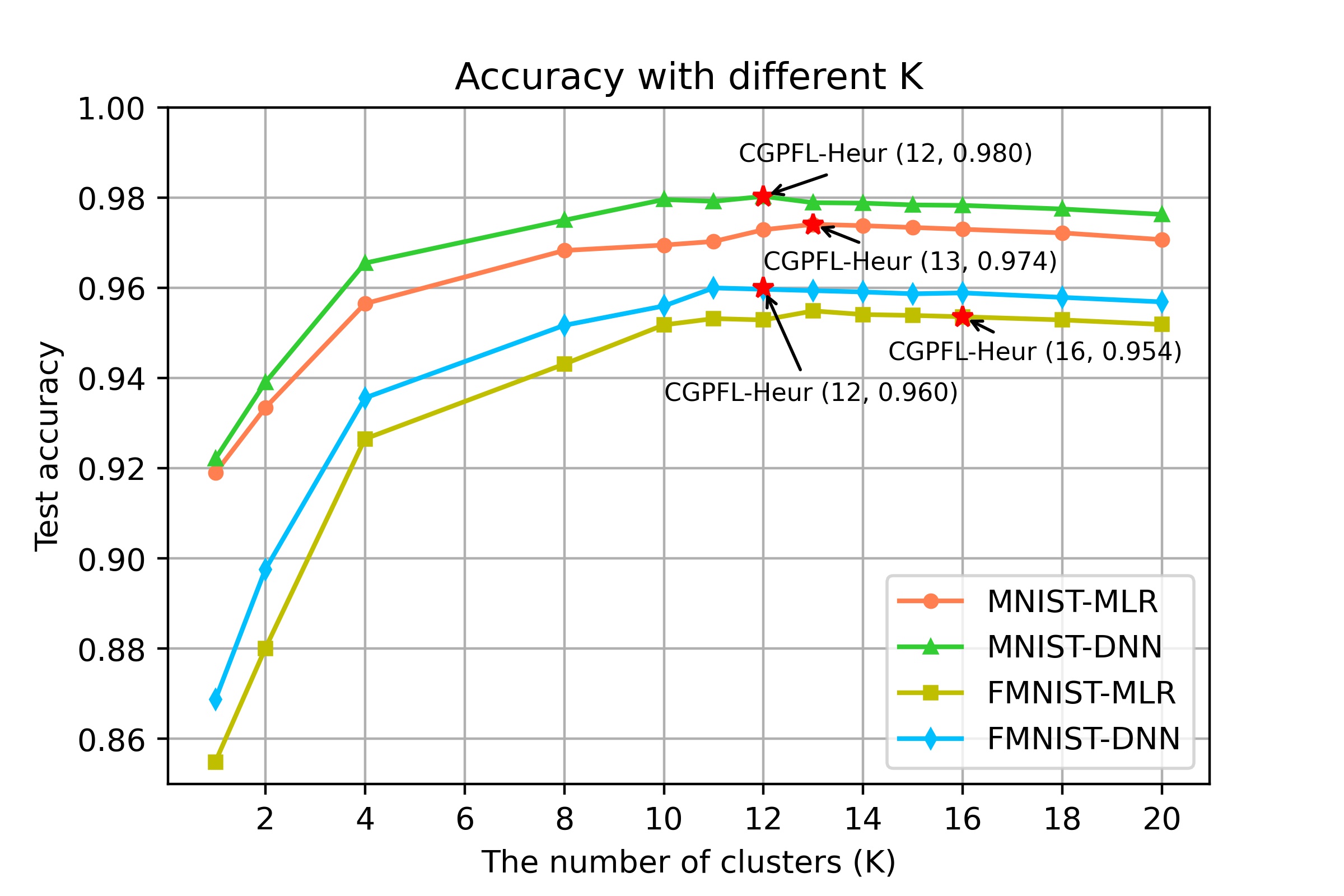}
    \end{minipage}
  }
  % \subfigure[\textit{CGPFL-Heur} versus \textit{CGPFL}]{
  \subfigure[\textit{CGPFL-Heur}]{
    \begin{minipage}{0.22\textwidth}
    \centering
    \label{acc-heur-optimal}
    \includegraphics[width=1.1\textwidth]{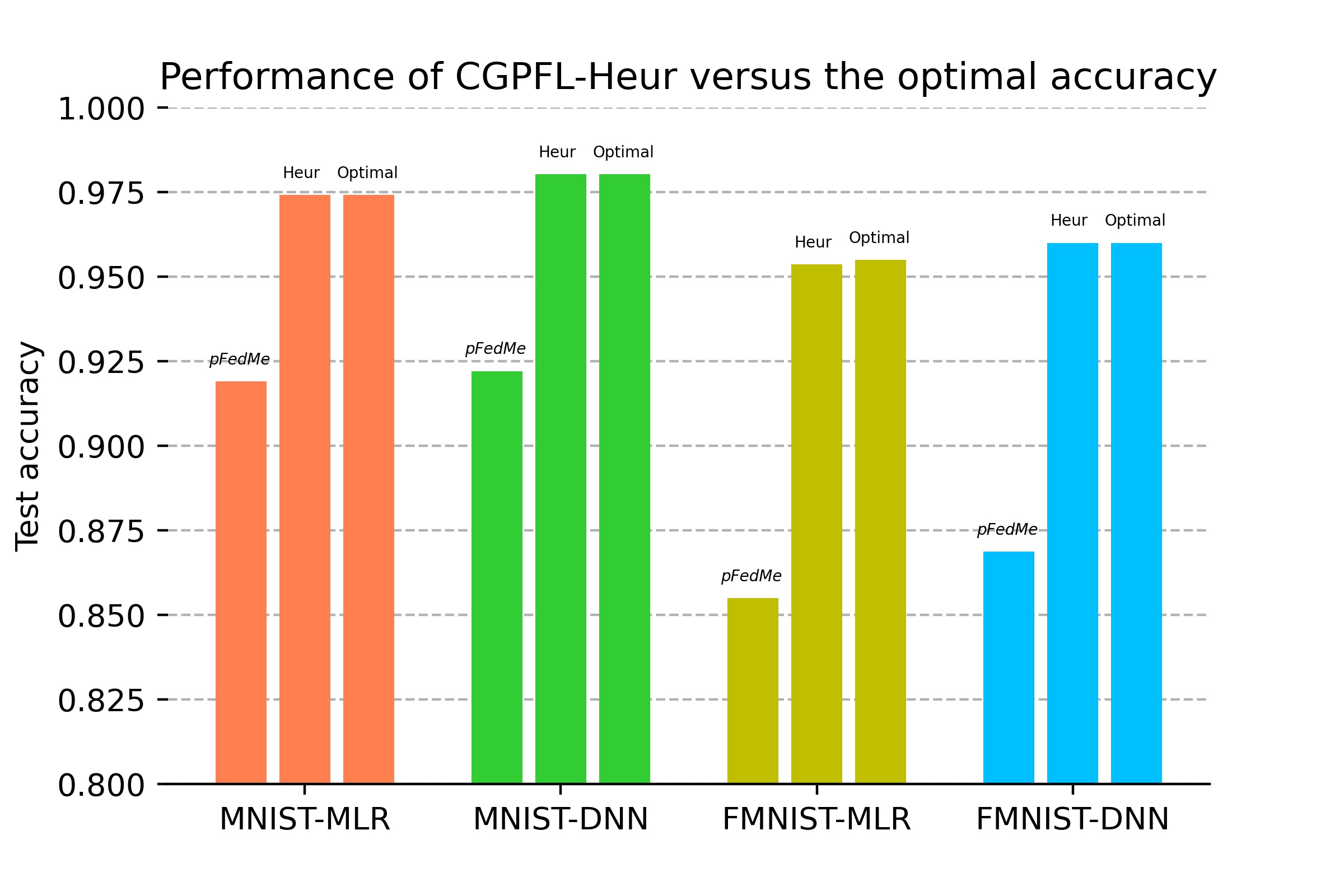}
    \end{minipage}
  }
  \caption{Further evaluation on \textit{CGPFL-Heur} on MNIST and FMNIST datasets}
  \label{fig:cgpfl-heur}
  %\Description{}
\vspace{-0.27cm}
\end{figure}

To further evaluate the performance of \textit{CGPFL-Heur}, on the one hand, we conduct the \textit{CGPFL} training with different number of contexts (i.e., $K$) varying form $1$ to $N/2$ on MINST and FMNIST, respectively. 
Specifically, we set the maximal value of $K$ no more than $N/2$ to avoid overfitting. By collating the model accuracy with different $K$, we can find out the optimal $K$ which corresponds to the optimal personalization-generalization trade-off in \textit{CGPFL}. The results are demonstrated in Figure~\ref{acc-optimal-K}. 
On the other hand, we conduct the \textit{CGPFL-Heur} training with an appropriate $\mu$ and keep other parameters same as that of the above evaluation. As shown in Figure~\ref{acc-optimal-K}, we distinguish the results of \textit{CGPFL-Heur} using red-star points. 
Besides, we make comparisons between the performance of a state-of-the-art PFL algorithm, \textit{pFedMe}~\cite{dinh20pFedMe} with our proposed \textit{CGPFL} and \textit{CGPFL-Heur} in Figure~\ref{acc-heur-optimal}. The results in Figure~\ref{acc-optimal-K} and Figure~\ref{acc-heur-optimal} demonstrate that our designed heuristic algorithm \textit{CGPFL-Heur} can effectively reach a near-optimal trade-off and consequently achieve the near-optimal model accuracy.

\section{Conclusion}
In this paper, we propose a novel personalized federated learning framework, dubbed \textit{CGPFL}, to handle the challenge of statistical heterogeneity (Non-I.I.D.), especially contextual heterogeneity in the federated setting. 
To the best of our knowledge, we are the first to propose the concept of contextualized generalization (CG) for personalized federated learning and further formulate it to a bi-level optimization problem that is solved effectively. 
Our method provides fine-grained generalization for personalized models which can prompt higher test accuracy and facilitate faster model convergence. 
Experimental results on real-world datasets demonstrate the effectiveness of our method over the state-of-the-art works.

% References

% \newpage

\section*{Acknowledgments}
This research was supported by fundings from the Key-Area Research and Development Program of Guangdong Province (No. 2021B0101400003), Hong Kong RGC Research Impact Fund (No. R5060-19), General Research Fund (No. 152221/19E, 152203/20E, and 152244/21E), the National Natural Science Foundation of China (61872310), and Shenzhen Science and Technology Innovation Commission (JCYJ20200109142008673).

\bibliographystyle{named}
\bibliography{CGPFL-references}

\newpage
% Appendix
\appendix

\section{Analysis of Convergence}

\subsection{The Iterates of Model Parameters}
The local update is given as follows:
\begin{displaymath}
  {\omega}_{i,r+1}^t = {\omega}_{i,r}^t - \beta\nabla G_i({\omega}_{i,r}^t) = {\omega}_{i,r}^t - \beta\underbrace{\frac{2}{N}({\omega}_{i,r}^t - \tilde{\theta}_i({\omega}_{i,r}^t))}_{:= h_{i,r}^t},
\end{displaymath}
Suming the local iterats, we can get
\begin{displaymath}
  \beta\sum_{r=0}^{R-1} h_{i,r}^t = \sum_{r=0}^{R-1}({\omega}_{i,r}^t - {\omega}_{i,r+1}^t) = {\omega}_{i,0}^t - {\omega}_{i,R}^t.
\end{displaymath}
According to the algorithm, we have
\begin{displaymath}
  {\Omega}_K^{t+1} - {\Omega}_K^t = -\alpha({\Omega}_K^t - {\Omega}_{I,R}^t P^{t+1}).
\end{displaymath}
Therefore, we can get the model parameters of the global models as follows:
\begin{displaymath}
\begin{aligned}
  {\Omega}_K^{t+1} &= (1-\alpha){\Omega}_K^t + \alpha{\Omega}_{I,R}^t P^{t+1}
  \\&= (1-\alpha){\Omega}_K^t + \alpha({\Omega}_{I,0}^t - \beta R
  \underbrace{\frac{1}{R}\sum_{r=0}^{R-1}H_{I,r}^t}_{:= H_I^t})P^{t+1}
  \\&= (1-\alpha){\Omega}_K^t + \alpha{\Omega}_K^t J^t P^{t+1} - 
  \underbrace{\alpha\beta R}_{:= \hat{\alpha}} H_I^t P^{t+1}
  \\&= (1-\alpha){\Omega}_K^t + \alpha{\Omega}_K^t J^t P^t Q^t - \hat{\alpha} H_I^t P^{t+1}
  \\&= (1-\alpha){\Omega}_K^t + \alpha{\Omega}_K^t Q^t - \hat{\alpha} H_I^t P^{t+1}
  \\&= {\Omega}_K^t[(1 - \alpha)I_K + \alpha Q^t] - \hat{\alpha} H_I^t P^t Q^t
\end{aligned}
\end{displaymath}
That is
\begin{equation}
  \label{param_update}
  {\Omega}_K^t - {\Omega}_K^{t+1} = \alpha{\Omega}_K^t(I_K - Q^t) + \hat{\alpha} H_I^t P^{t+1}.
\end{equation}
It's noted that
\begin{displaymath}
  G_k({\omega}_k^t) := \big[G_K({\Omega}_K^t)\big]_k,
\end{displaymath}
where $[G_K({\Omega}_K^t)]_k$ denotes the $k$-th element of the row vector $G_K({\Omega}_K^t)$.

% \subsection{Review of Useful Propositions and Lemmas}
\noindent\textbf{Assumption 3} (bounded parameters and gradients) 
{\itshape The generalized model parameters ${\Omega}_K^t$ 
and the gradients $\nabla G_K({\Omega}_K^t)$ 
are upper bounded by ${\rho}_{\Omega}$ 
and ${\rho}_g$, respectively.}
\begin{small}
\begin{gather}
  {\big\lVert {\Omega}_K^t \big\rVert}^2 \leq {\rho}_{\Omega}^2 \qquad 
  \text{and} 
  \qquad {\big\lVert \nabla G_K({\Omega}_K^t) \big\rVert}^2 \leq {\rho}_g^2 \text{, } \quad \forall t
\end{gather}
\end{small}where ${\rho}_{\Omega}$ and ${\rho}_{g}$ are finite non-negative constants.

\subsection{Convergence Rate}

{\itshape Proof:}
\begin{flalign*}
&\mathbb{E}\Big[\sum_{k=1}^{K}G_k({\omega}_k^{t+1}) - \sum_{k=1}^{K}G_k({\omega}_k^t) \Big] &
\\&= \mathbb{E}\Big[\sum_{k=1}^{K}[G_I({\Omega}_K^{t+1})P^{t+1}]_k - \sum_{k=1}^{K}[G_I({\Omega}_K^t)P^t]_k \Big] &
\\&= \mathbb{E}\Big[ \sum_{k=1}^{K}[G_I({\Omega}_K^{t+1})P^{t+1} - G_I({\Omega}_K^t)P^t]_k \Big] &
% \\&= \mathbb{E}\Big[ \sum_{k=1}^{K}[G_I({\Omega}_K^{t+1})P^{t+1} - G_I({\Omega}_K^t)P^t]_k \Big] &
\\&= \mathbb{E}\Big[ \sum_{k=1}^{K}\big[\big(G_I({\Omega}_K^{t+1}) - G_I({\Omega}_K^t) \big)P^t\big]_k &
\\&\quad+ \sum_{k=1}^{K}\big[G_I({\Omega}_K^{t+1})P^t(Q^t - I_K) \big]_k \Big] &
\\&\leq \underbrace{\mathbb{E}\Big[\big\langle \nabla G_K({\Omega}_K^t), {\Omega}_K^{t+1} - {\Omega}_K^t \big\rangle \Big] + \frac{L_G}{2}\mathbb{E}\Big[{\big\lVert {\Omega}_K^{t+1} - {\Omega}_K^t \big\rVert}^2 \Big]}_{\mathbf{A}} &
\\&\quad+ \underbrace{\mathbb{E}\Big[\sum_{k=1}^{K}\big[G_I({\Omega}_K^{t+1})P^t(Q^t - I_K) \big]_k \Big]}_{\mathbf{B}}, &
\end{flalign*}
where we assume that $L_G := \mathop{max}_{k \in [K]}L_{G_k}$.
We first deal with the part $\mathbf{A}$ in above inequation. According to the above derivation, we have
\begin{flalign*}
\mathbf{A} &= \mathbb{E}\Big[\big\langle \nabla G_K({\Omega}_K^t), {\Omega}_K^{t+1} - {\Omega}_K^t \big\rangle \Big] + \frac{L_G}{2}\mathbb{E}\Big[{\big\lVert {\Omega}_K^{t+1} - {\Omega}_K^t \big\rVert}^2 \Big] &
\\&= -\hat{\alpha}\mathbb{E}\Big[\big\langle \nabla G_K({\Omega}_K^t), \frac{1}{\hat{\alpha}}\big({\Omega}_K^{t} - {\Omega}_K^{t+1}\big) - \nabla G_K({\Omega}_K^t) & 
\\&\quad+ \nabla G_K({\Omega}_K^t) \big\rangle \Big] + \frac{L_G}{2}\mathbb{E}\Big[{\big\lVert {\Omega}_K^{t+1} - {\Omega}_K^t \big\rVert}^2 \Big] &
\\&= -\hat{\alpha} \mathbb{E}\Big[{\big\lVert \nabla G_K({\Omega}_K^t)\big\rVert}^2\Big] + \frac{L_G}{2}\mathbb{E}\Big[{\big\lVert {\Omega}_K^{t+1} - {\Omega}_K^t \big\rVert}^2 \Big] &
\\&\quad-\hat{\alpha}\mathbb{E}\Big[\big\langle \nabla G_K({\Omega}_K^t), \frac{1}{\hat{\alpha}}\big({\Omega}_K^{t} - {\Omega}_K^{t+1}\big) - \nabla G_K({\Omega}_K^t) \big\rangle \Big] &
\end{flalign*}
\begin{flalign*}
&\leq -\hat{\alpha} \mathbb{E}\Big[{\big\lVert \nabla G_K({\Omega}_K^t)\big\rVert}^2\Big] + \frac{\hat{\alpha}}{2}\mathbb{E}\Big[{\big\lVert \nabla G_K({\Omega}_K^t)\big\rVert}^2\Big] &
\\&\quad+ \frac{\hat{\alpha}}{2}\mathbb{E}\Big[{\big\lVert \frac{1}{\hat{\alpha}}\big({\Omega}_K^{t} - {\Omega}_K^{t+1}\big) - \nabla G_K({\Omega}_K^t)\big\rVert}^2\Big] &
\\& \quad + \frac{L_G}{2}\mathbb{E}\Big[{\big\lVert {\Omega}_K^{t+1} - {\Omega}_K^t \big\rVert}^2 \Big] &
\\&= -\frac{\hat{\alpha}}{2}\mathbb{E}\Big[{\big\lVert \nabla G_K({\Omega}_K^t)\big\rVert}^2\Big] + \underbrace{\frac{L_G}{2}\mathbb{E}\Big[{\big\lVert {\Omega}_K^{t+1} - {\Omega}_K^t \big\rVert}^2 \Big]}_{\mathbf{A_1}} &
\\&\quad+ \underbrace{\frac{\hat{\alpha}}{2}\mathbb{E}\Big[{\big\lVert \frac{1}{\hat{\alpha}}\big({\Omega}_K^{t} - {\Omega}_K^{t+1}\big) - \nabla G_K({\Omega}_K^t)\big\rVert}^2\Big]}_{\mathbf{A_2}} &
\end{flalign*}
Plugging equation ~\eqref{param_update} into above inequation, we can get
\begin{flalign*}
\mathbf{A_1} &= \frac{L_G}{2}\mathbb{E}\Big[{\big\lVert \alpha{\Omega}_K^t(I_K - Q^t) + \hat{\alpha} H_I^t P^{t+1} \big\rVert}^2\Big] &
\\&= \frac{L_G}{2}\mathbb{E}\Big[{\big\lVert \alpha{\Omega}_K^t(I_K - Q^t)+\hat{\alpha} H_I^t P^{t+1}} &
\\&\quad{- \hat{\alpha}\nabla G_I({\Omega}_{I,0}^t)P^{t+1} + \hat{\alpha}\nabla G_I({\Omega}_{I,0}^t)P^{t}Q^t \big\rVert}^2\Big] &
\\& \leq \frac{3{\alpha}^2L_G}{2}\mathbb{E}\Big[{\big\lVert {\Omega}_K^t(I_K-Q^t)\big\rVert}^2\Big] &
\\&\quad+ \frac{3{\hat{\alpha}}^2L_G}{2}\mathbb{E}\Big[{\big\lVert \nabla G_K({\Omega}_K^t)Q^t \big\rVert}^2\Big] &
\\& \quad + \frac{3{\hat{\alpha}}^2L_G}{2}\mathbb{E}\Big[{\big\lVert \big(H_I^t - \nabla G_I({\Omega}_{I,0}^t)\big)P^{t+1}\big\rVert}^2 \Big] &
\end{flalign*}
and
\begin{small}
\begin{flalign*}
\mathbf{A_2} &= \frac{\hat{\alpha}}{2}\mathbb{E}\Big[{\big\lVert \frac{\alpha}{\hat{\alpha}}{\Omega}_K^t(I_K - Q^t) + H_I^t P^{t+1} - \nabla G_I({\Omega}_{I,0}^t)P^{t+1}} &
\\&\quad{+ \nabla G_I({\Omega}_{I,0}^t)P^{t}Q^t - \nabla G_K({\Omega}_K^t)\big\rVert}^2\Big] &
\\& \leq \frac{3{\alpha}^2}{2\hat{\alpha}}\mathbb{E}\Big[{\big\lVert {\Omega}_K^t(I_K-Q^t)\big\rVert}^2\Big] + \frac{3\hat{\alpha}}{2}\mathbb{E}\Big[{\big\lVert \nabla G_K({\Omega}_{K}^t)(I_K - Q^t)\big\rVert}^2\Big] &
\\&\quad+ \frac{3\hat{\alpha}}{2}\mathbb{E}\Big[{\big\lVert \big(H_I^t - \nabla G_I({\Omega}_{I,0}^t)\big)P^{t+1}\big\rVert}^2 \Big] &
\end{flalign*}
\end{small}\noindent\textbf{Proposition 2} {\itshape For any vector $x_i \in \mathbb{R}^d, i = 1,2,\dots,M$, according to Jensen's inequality, we have}
\begin{displaymath}
{\Big\lVert \sum_{i=1}^M x_i \Big\rVert}^2 \leq M\sum_{i=1}^M {\lVert x_i \rVert}^2.
\end{displaymath}
{\itshape And because the real function $\varphi(y) = y^2, y \in 
\mathbb{R}$ is convex, if some constants satisfy that ${\lambda}_i \geq 0, \forall i = 1,2,\dots,M$, and $\sum_{i=1}^M {\lambda}_i = 1$, we have}
\begin{displaymath}
{\Big\lVert \sum_{i=1}^M {\lambda}_i y_i \Big\rVert}^2 \leq \sum_{i=1}^M {\lambda}_i{\lVert y_i \rVert}^2.
\end{displaymath}
\noindent\textbf{Lemma 1} {\itshape We can obtain that $\mathbb{E}\Big[{\big\lVert XP^{t+1} \big\rVert}^2\Big]\leq\mathbb{E}\Big[{\big\lVert X \big\rVert}^2\Big]$, and $\mathbb{E}\Big[{\big\lVert YQ^t \big\rVert}^2\Big]\leq\mathbb{E}\Big[{\big\lVert Y \big\rVert}^2\Big]$ for any matrices $X \in \mathbb{R}^{d\times N}$ and $Y \in \mathbb{R}^{d\times K}$, as long as the $P^{t+1}$ and $Q^t$ satisfy that $\sum_{i=1}^{N}P_{i,k}^{t+1} = 1$, $\sum_{j=1}^{K}Q_{j,k}^{t} = 1 \forall k, t$, and $\sum_{k=1}^{K}Q_{j,k}^{t} = 1, \forall j, t$}. Especialy in this paper, we have 
$P_{i,k}^{t+1}=
\begin{cases}
\frac{1}{\lvert C_k \rvert}, & \mbox{if } i \in C_k \\
0, & \mbox{otherwise}
\end{cases}$.

\noindent{\itshape Proof:}
\begin{flalign*}
\mathbb{E}\Big[{\big\lVert XP^{t+1} \big\rVert}^2\Big] &= \sum_{l=1}^{d}\sum_{k=1}^{K}\big[(XP^{t+1})_{l,k}\big]^2 &
\\&= \sum_{l=1}^{d}\sum_{k=1}^{K}\Big[\sum_{i=1}^N X_{l,i}P_{i,k}^{t+1}\Big]^2 &
\\&\leq \sum_{l=1}^{d}\sum_{k=1}^{K}\sum_{i=1}^N {X_{l,i}}^2 P_{i,k}^{t+1} &
\\&= \sum_{l=1}^{d}\sum_{i=1}^N\sum_{k=1}^{K} {X_{l,i}}^2 P_{i,k}^{t+1} &
\\&= \sum_{l=1}^{d}\sum_{i=1}^N {X_{l,i}}^2 \sum_{k=1}^{K} P_{i,k}^{t+1} &
\\&\leq \sum_{l=1}^{d}\sum_{i=1}^N {X_{l,i}}^2 = \mathbb{E}\big[{\lVert X \rVert}^2\big] &
\end{flalign*}
Similiarly,
\begin{flalign*}
\mathbb{E}\big[{\lVert YQ^{t} \rVert}^2\big] &= \sum_{l=1}^{d}\sum_{k=1}^{K}\big[(YQ^{t})_{l,k}\big]^2 &
\\&= \sum_{l=1}^{d}\sum_{k=1}^{K}\Big[\sum_{j=1}^K Y_{l,j}Q_{j,k}^{t}\Big]^2 &
\\&\leq \sum_{l=1}^{d}\sum_{k=1}^{K}\sum_{j=1}^K {Y_{l,j}}^2 Q_{j,k}^{t} = \sum_{l=1}^{d}\sum_{k=1}^K\sum_{j=1}^{K} {Y_{l,j}}^2 Q_{j,k}^{t} &
\\&= \sum_{l=1}^{d}\sum_{j=1}^N {Y_{l,j}}^2 \sum_{k=1}^{K} Q_{j,k}^{t} &
\\&= \sum_{l=1}^{d}\sum_{j=1}^N {Y_{l,j}}^2 = \mathbb{E}\big[{\lVert Y \rVert}^2\big] &
\end{flalign*}

In the next part, we will first cope with $\mathbb{E}\Big[{\big\lVert H_I^t - \nabla G_I({\Omega}_{I,0}^t)\big\rVert}^2 \Big]$

\begin{flalign*}
&\mathbb{E}\Big[{\big\lVert \big(G_I^t - \nabla F_I({\Omega}_{I,0}^t)\big)P^{t+1}\big\rVert}^2 \Big] &
\\&= \mathbb{E}\bigg[{\Big\lVert \frac{1}{R}\sum_{r=0}^{R-1}\big(H_{I,r}^t - \nabla G_I({\Omega}_{I,0}^t) \big)P^{t+1} \Big\rVert}^2 \bigg] &
\\&\leq \frac{1}{R}\sum_{r=0}^{R-1}\mathbb{E}\Big[{\big\lVert \big(H_{I,r}^t - \nabla G_I({\Omega}_{I,0}^t)\big)P^{t+1}\big\rVert}^2 \Big] &
\end{flalign*}
\begin{flalign*}
&= \frac{1}{R}\sum_{r=0}^{R-1}\mathbb{E}\Big[{\big\lVert \big(H_{I,r}^t - \nabla G_I({\Omega}_{I,r}^t) + \nabla G_I({\Omega}_{I,r}^t)} &
\\&\quad{- \nabla G_I({\Omega}_{I,0}^t)\big)P^{t+1}\big\rVert}^2 \Big] &
\\&\leq \frac{2}{R}\sum_{r=0}^{R-1}\mathbb{E}\Big[{\big\lVert \big(H_{I,r}^t - \nabla G_I({\Omega}_{I,r}^t)\big)P^{t+1}\big\rVert}^2 \Big] &
\\&\quad+ \frac{2}{R}\sum_{r=0}^{R-1}\mathbb{E}\Big[{\big\lVert \big(\nabla G_I({\Omega}_{I,r}^t) - \nabla G_I({\Omega}_{I,0}^t)\big)P^{t+1}\big\rVert}^2 \Big] &
\\&\leq \frac{2}{R}\sum_{r=0}^{R-1}\mathbb{E}\Big[{\big\lVert H_{I,r}^t - \nabla G_I({\Omega}_{I,r}^t) \big\rVert}^2 \Big] &
\\&\quad+ \frac{2}{R}\sum_{r=0}^{R-1}\mathbb{E}\Big[{\big\lVert \big(\nabla G_I({\Omega}_{I,r}^t) - \nabla G_I({\Omega}_{I,0}^t)\big)P^{t+1}\big\rVert}^2 \Big] &
\\&\leq \frac{2}{R}\sum_{r=0}^{R-1}\mathbb{E}\Big[{\big\lVert \frac{2}{N}(\widetilde{\Theta}_i({\Omega}_{I,r}^t) - \widehat{\Theta}_i({\Omega}_{I,r}^t)) \big\rVert}^2\Big] &
\\&\quad+ \frac{2{L_G}^2}{R}\sum_{r=0}^{R-1}\mathbb{E}\Big[{\big\lVert \big({\Omega}_{I,r}^t - {\Omega}_{I,0}^t\big)P^{t+1} \big\rVert}^2\Big] &
\\&\leq \frac{8}{N}{\delta}^2 + \frac{2{L_G}^2}{R}\sum_{r=0}^{R-1}\mathbb{E}\Big[{\big\lVert \big({\Omega}_{I,r}^t - {\Omega}_{I,0}^t\big)P^{t+1} \big\rVert}^2\Big] &
\end{flalign*}
Because
\begin{small}
\begin{flalign*}
&\mathbb{E}\Big[{\big\lVert \big({\Omega}_{I,r}^t - {\Omega}_{I,0}^t\big)P^{t+1} \big\rVert}^2\Big] &
\\&=  \mathbb{E}\Big[{\big\lVert \big({\Omega}_{I,r-1}^t - {\Omega}_{I,0}^t - \beta H_{I,r-1}^t\big)P^{t+1} \big\rVert}^2\Big] &
\\&= \mathbb{E}\Big[{\big\lVert \big({\Omega}_{I,r-1}^t - {\Omega}_{I,0}^t - \beta\nabla G_I({\Omega}_{I,0}^t) + \beta\nabla G_I({\Omega}_{I,0}^t)} &
\\&\quad{- \beta H_{I,r-1}^t\big)P^{t+1} \big\rVert}^2\Big] &
\\&\leq (1+\frac{1}{R})\mathbb{E}\Big[{\big\lVert \big({\Omega}_{I,r-1}^t - {\Omega}_{I,0}^t - \beta\nabla G_I({\Omega}_{I,0}^t)\big)P^{t+1} \big\rVert}^2\Big] &
\\&\quad+ (1+R){\beta}^2\mathbb{E}\Big[{\big\lVert \big(\nabla G_I({\Omega}_{I,0}^t) - H_{I,r-1}^t\big)P^{t+1} \big\rVert}^2\Big] &
\\&\leq (1+\frac{1}{R})(1+\frac{1}{2R})\mathbb{E}\Big[{\big\lVert \big({\Omega}_{I,r-1}^t - {\Omega}_{I,0}^t\big)P^{t+1} \big\rVert}^2\Big] &
\\&\quad+ (1+\frac{1}{R})(1+2R){\beta}^2\mathbb{E}\Big[{\big\lVert \nabla G_I({\Omega}_{I,0}^t)P^t Q^t \big\rVert}^2\Big] &
\\&\quad + {\beta}^2(1+R)\Big( \frac{8}{N}{\delta}^2 + 2{L_G}^2\mathbb{E}\Big[{\big\lVert \big({\Omega}_{I,r-1}^t - {\Omega}_{I,0}^t\big)P^{t+1} \big\rVert}^2\Big] \Big) &
\\&= (1+\frac{1}{R})\big(1+\frac{1}{2R}+2(1+R){\beta}^2{L_G}^2\big)\mathbb{E}\Big[{\big\lVert \big({\Omega}_{I,r-1}^t - {\Omega}_{I,0}^t\big)P^{t+1} \big\rVert}^2\Big] &
\\&\quad+ (1+\frac{1}{R})(1+2R){\beta}^2\mathbb{E}\Big[{\big\lVert \nabla G_K({\Omega}_{K}^t)Q^t \big\rVert}^2\Big] + \frac{8(1+R){\beta}^2}{N}{\delta}^2 &
\\&\leq (1+\frac{1}{R})^2\mathbb{E}\Big[{\big\lVert \big({\Omega}_{I,r-1}^t - {\Omega}_{I,0}^t\big)P^{t+1} \big\rVert}^2\Big] &
\\&\quad+ (1+\frac{1}{R})(1+2R){\beta}^2\mathbb{E}\Big[{\big\lVert \nabla G_K({\Omega}_{K}^t) \big\rVert}^2\Big] + \frac{8(1+R){\beta}^2}{N}{\delta}^2 &
\end{flalign*}
\end{small}with ${\beta}^2 \leq \frac{1}{4R(1+R){L_G}^2}$, which implies 
that $2(1+R){\beta}^2{L_G}^2 \leq \frac{1}{2R}$. By unrolling the above result recursively, we can get

\begin{flalign*}
&\mathbb{E}\Big[{\big\lVert \big({\Omega}_{I,r}^t - {\Omega}_{I,0}^t\big)P^{t+1} \big\rVert}^2\Big] &
\\&\leq \Big\{ (1+\frac{1}{R})(1+2R){\beta}^2\mathbb{E}\Big[{\big\lVert \nabla G_K({\Omega}_{K}^t) \big\rVert}^2\Big] &
\\&\quad+ \frac{8(1+R){\beta}^2}{N}{\delta}^2 \Big\}\sum_{\hat{r}=0}^{r-2}\big(1+\frac{1}{R}\big)^{2\hat{r}} &
\\&\leq \Big\{ (1+\frac{1}{R})(1+2R){\beta}^2\mathbb{E}\Big[{\big\lVert \nabla G_K({\Omega}_{K}^t) \big\rVert}^2\Big] &
\\&\quad+ \frac{8(1+R){\beta}^2}{N}{\delta}^2 \Big\} \frac{(1+\frac{1}{R})^{2(r-1)}-1}{(1+\frac{1}{R})^2-1} &
\\&\leq \Big\{ (1+\frac{1}{R})(1+2R){\beta}^2\mathbb{E}\Big[{\big\lVert \nabla G_K({\Omega}_{K}^t) \big\rVert}^2\Big] &
\\&\quad+ \frac{8(1+R){\beta}^2}{N}{\delta}^2 \Big\} \frac{(1+\frac{1}{R})^{2(r-1)}}{(1+\frac{1}{R})^2-1} &
\end{flalign*}
and then 
\begin{flalign*}
&\mathbb{E}\Big[{\big\lVert \big(H_I^t - \nabla G_I({\Omega}_{I,0}^t)\big)P^{t+1}\big\rVert}^2 \Big] &
\\&\leq \frac{8}{N}{\delta}^2 + \frac{2{L_G}^2}{R}\sum_{r=0}^{R-1}\mathbb{E}\Big[{\big\lVert \big({\Omega}_{I,r}^t - {\Omega}_{I,0}^t\big)P^{t+1} \big\rVert}^2\Big]  &
\\&\leq \frac{8}{N}{\delta}^2 + \frac{2{\beta}^2{L_G}^2}{R}\Big\{ (1+\frac{1}{R})(1+2R)\mathbb{E}\Big[{\big\lVert \nabla G_K({\Omega}_{K}^t) \big\rVert}^2\Big] &
\\&\quad+ \frac{8(1+R)}{N}{\delta}^2 \Big\}\sum_{r=0}^{R-1}\frac{(1+\frac{1}{R})^{2(r-1)}}{(1+\frac{1}{R})^2-1} &
\\&\leq \frac{8}{N}{\delta}^2 + \frac{2{\beta}^2{L_G}^2}{R}\Big\{ (1+\frac{1}{R})(1+2R)\mathbb{E}\Big[{\big\lVert \nabla G_K({\Omega}_{K}^t) \big\rVert}^2\Big] &
\\&\quad+ \frac{8(1+R)}{N}{\delta}^2 \Big\}\frac{(1+\frac{1}{R})^{2R}-1}{(1+\frac{1}{R})^2-1} &
\\&\leq \frac{8}{N}{\delta}^2 + \frac{2{\beta}^2{L_G}^2}{R}\Big\{ (1+\frac{1}{R})(1+2R)\mathbb{E}\Big[{\big\lVert \nabla G_K({\Omega}_{K}^t) \big\rVert}^2\Big] &
\\&\quad+ \frac{8(1+R)}{N}{\delta}^2 \Big\}\frac{e^2-1}{(1+\frac{1}{R})^2-1} &
\\&\leq \frac{8}{N}{\delta}^2 + \frac{2{\beta}^2{L_G}^2}{R}\Big\{ (1+\frac{1}{R})(1+2R)\mathbb{E}\Big[{\big\lVert \nabla G_K({\Omega}_{K}^t) \big\rVert}^2\Big] &
\\&\quad+ \frac{8(1+R)}{N}{\delta}^2 \Big\}\frac{8R^2}{1+2R} &
\\&= \frac{8}{N}{\delta}^2 + \frac{128R(1+R){\beta}^2{L_G}^2{\delta}^2}{(1+2R)N} &
\\&\quad+ 16(1+R){\beta}^2{L_G}^2\mathbb{E}\Big[{\big\lVert \nabla G_K({\Omega}_{K}^t) \big\rVert}^2\Big] &
\\& \leq \frac{8}{N}{\delta}^2 + \frac{128R{\beta}^2{L_G}^2{\delta}^2}{N} + 32R{\beta}^2{L_G}^2\mathbb{E}\Big[{\big\lVert \nabla G_K({\Omega}_{K}^t) \big\rVert}^2\Big] &
\end{flalign*}

Therefore, we can obtain
\begin{flalign*}
\mathbf{A} &= -\frac{\hat{\alpha}}{2}\mathbb{E}\Big[{\big\lVert \nabla G_K({\Omega}_K^t)\big\rVert}^2\Big] + \mathbf{A_1} + \mathbf{A_2} &
\\&\leq \big(-\frac{\hat{\alpha}}{2} + \frac{3{\hat{\alpha}}^2 L_G}{2}\big)\mathbb{E}\Big[{\big\lVert \nabla G_K({\Omega}_K^t)\big\rVert}^2\Big] &
\\&\quad+ \big(\frac{3{\hat{\alpha}}^2 L_G}{2} + \frac{3\hat{\alpha}}{2}\big)\mathbb{E}\Big[{\big\lVert \big(H_I^t - \nabla G_I({\Omega}_{I,0}^t)\big)P^{t+1}\big\rVert}^2 \Big] &
\\&\quad+ \big(\frac{3{\alpha}^2 L_G}{2} + \frac{3{\alpha}^2}{2\hat{\alpha}}\big)\underbrace{\mathbb{E}\Big[{\big\lVert {\Omega}_K^t(I_K-Q^t)\big\rVert}^2\Big]}_{\mathbf{B_1}} &
\\&\quad+ \frac{3\hat{\alpha}}{2}\underbrace{\mathbb{E}\Big[{\big\lVert \nabla G_K({\Omega}_{K}^t)(I_K - Q^t)\big\rVert}^2\Big]}_{\mathbf{B_2}} &
\end{flalign*}
That is,
\begin{small}
\begin{flalign*}
\mathbf{A} &\leq -\frac{\hat{\alpha}}{2}(1 - 3{\hat{\alpha}} L_G)\mathbb{E}\Big[{\big\lVert \nabla G_K({\Omega}_K^t)\big\rVert}^2\Big] &
\\&\quad+ \frac{3\hat{\alpha}}{2}(1 + {\hat{\alpha}} L_G)\mathbb{E}\Big[{\big\lVert \big(H_I^t - \nabla G_I({\Omega}_{I,0}^t)\big)P^{t+1}\big\rVert}^2 \Big] &
\\&\quad+ \frac{3{\alpha}^2}{2}\big(L_G + \frac{1}{\hat{\alpha}}\big)\mathbf{B_1} + \frac{3\hat{\alpha}}{2}\mathbf{B_2} &
\\&\leq -\frac{\hat{\alpha}}{2}(1 - 3{\hat{\alpha}} L_G)\mathbb{E}\Big[{\big\lVert \nabla G_K({\Omega}_K^t)\big\rVert}^2\Big] &
\\&\quad+ \frac{3{\alpha}^2}{2}\big(L_G + \frac{1}{\hat{\alpha}}\big)\mathbf{B_1} + \frac{3\hat{\alpha}}{2}\mathbf{B_2} &
\\&\quad+ \frac{3\hat{\alpha}}{2}(1 + {\hat{\alpha}} L_G) \bigg\{\frac{8}{N}{\delta}^2 + \frac{128R{\beta}^2{L_G}^2{\delta}^2}{N} &
\\&\quad+ 32R{\beta}^2{L_G}^2\mathbb{E}\Big[{\big\lVert \nabla G_K({\Omega}_{K}^t) \big\rVert}^2\Big] \bigg\} &
\\&= -\frac{\hat{\alpha}}{2}\big(1 - 3{\hat{\alpha}} L_G - 96R{\beta}^2 {L_G}^2(1 + {\hat{\alpha}} L_G)\big)\mathbb{E}\Big[{\big\lVert \nabla G_K({\Omega}_K^t)\big\rVert}^2\Big] &
\\&\quad+ \frac{3{\alpha}^2}{2}\big(L_G + \frac{1}{\hat{\alpha}}\big)\mathbf{B_1} + \frac{3\hat{\alpha}}{2}\mathbf{B_2} &
\\&\quad+ \frac{192{\hat{\alpha}}^3{\delta}^2{L_G}^2(1+{\hat{\alpha}} L_G)}{NR{\alpha}^2} + \frac{12\hat{\alpha} (1 + {\hat{\alpha}} L_G){\delta}^2}{N} &
\\&= -\frac{\hat{\alpha}}{2}\underbrace{\big(1 - 3{\hat{\alpha}} L_G - 96R{\beta}^2 {L_G}^2(1 + {\hat{\alpha}} L_G)\big)}_{\geq \frac{1}{2} \text{ when } {\beta}^2{L_G}^2 \leq \frac{1}{416R^2} \text{ and } \alpha \leq 1}\mathbb{E}\Big[{\big\lVert \nabla G_K({\Omega}_K^t)\big\rVert}^2\Big] &
\\&\quad+ \frac{3{\alpha}^2}{2}\big(L_G + \frac{1}{\hat{\alpha}}\big)\mathbf{B_1} + \frac{3\hat{\alpha}}{2}\mathbf{B_2} &
\\&\quad+ \frac{192{\hat{\alpha}}^3{\delta}^2{L_G}^2(1+{\hat{\alpha}} L_G)}{NR{\alpha}^2} + \frac{12\hat{\alpha} (1 + {\hat{\alpha}} L_G){\delta}^2}{N} &
\\&\leq -\frac{\hat{\alpha}}{4}\mathbb{E}\Big[{\big\lVert \nabla G_K({\Omega}_K^t)\big\rVert}^2\Big] + \frac{192{\hat{\alpha}}^3{\delta}^2{L_G}^2(1+{\hat{\alpha}} L_G)}{NR{\alpha}^2} &
\\&\quad+ \frac{12\hat{\alpha} (1 + {\hat{\alpha}} L_G){\delta}^2}{N} + \frac{3{\alpha}^2}{2\hat{\alpha}}(\hat{\alpha} L_G + 1)\mathbf{B_1} + \frac{3\hat{\alpha}}{2}\mathbf{B_2} &
\\&\leq -\frac{\hat{\alpha}}{4}\mathbb{E}\Big[{\big\lVert \nabla G_K({\Omega}_K^t)\big\rVert}^2\Big] + \frac{2{\alpha}^2}{\hat{\alpha}}\mathbf{B_1} + \frac{3\hat{\alpha}}{2}\mathbf{B_2} &
\\&\quad+ \frac{208{\hat{\alpha}}^3{\delta}^2{L_G}^2}{NR{\alpha}^2} + \frac{13\hat{\alpha} {\delta}^2}{N} &
\end{flalign*}
\end{small}with ${\beta}^2{L_G}^2 \leq \frac{1}{416R^2} \leq \frac{1}{8R^2} \leq \frac{1}{4R(1+R)}$, $\forall R \geq 1$ and $\alpha \leq 1$.

Because of the above conditions we have
\begin{equation}
\label{constants1}
\hat{\alpha} L_G = R\alpha \beta L_G \leq \frac{R}{\sqrt{416R^2}} \leq \frac{1}{12},
\end{equation}
and
\begin{equation}
\label{constants2}
96R {\beta}^2{L_G}^2(1+\hat{\alpha}L_G) \leq \frac{96R}{416R^2}(1+\frac{1}{12}) = \frac{1}{4R} \leq \frac{1}{4}, \forall R \geq 1.
\end{equation}

\noindent Therefore,
\begin{equation*}
(1 - 3{\hat{\alpha}} L_G - 96R{\beta}^2 {L_G}^2(1 + {\hat{\alpha}} L_G)\big) \geq 1 - \frac{1}{4} - \frac{1}{4R} \geq \frac{1}{2}.
\end{equation*}

\noindent\textbf{Lemma 2} With Assumption 1 held, we can get
\begin{flalign*}
&\mathbf{(1)}\quad\mathop{lim}_{T \to \infty}\frac{1}{T}\sum_{t=0}^{T-1}\underbrace{\mathbb{E}\Big[{\big\lVert {\Omega}_K^{t}(Q^t - I_K) \big\rVert}^2 \Big]}_{\mathbf{B_1}} = 0 &
\\&\quad\quad\Leftrightarrow \mathop{lim}_{T \to \infty}{\big\lVert Q^T - I_K \big\rVert}^2 = 0 &
\end{flalign*}
and
\begin{flalign*}
&\quad \frac{1}{T}\sum_{t=0}^{T-1}\mathbb{E}\Big[{\big\lVert {\Omega}_K^{t}(Q^t - I_K) \big\rVert}^2 \Big] = \mathcal{O}(\frac{1}{T}). &
\end{flalign*}

\begin{flalign*}
&\mathbf{(2)}\quad\mathop{lim}_{T \to \infty}\frac{1}{T}\sum_{t=0}^{T-1}\underbrace{\mathbb{E}\Big[{\big\lVert \nabla G_K({\Omega}_K^t)(Q^t - I_K) \big\rVert}^2 \Big]}_{\mathbf{B_2}} = 0 &
\\&\quad\quad\Leftrightarrow \mathop{lim}_{T \to \infty}{\big\lVert Q^T - I_K \big\rVert}^2 = 0 &
\end{flalign*}
and
\begin{flalign*}
&\quad \frac{1}{T}\sum_{t=0}^{T-1}\mathbb{E}\Big[{\big\lVert \nabla G_K({\Omega}_K^t)(Q^t - I_K) \big\rVert}^2 \Big] = \mathcal{O}(\frac{1}{T}). &
\end{flalign*}

\noindent{\itshape Proof:} \textbf{(1)}

\noindent \textbf{\itshape 1a)}\quad \textbf{" $\Longrightarrow$ ":} \quad We have 
\begin{displaymath}
\mathop{lim}_{T \to \infty}\frac{1}{T}\sum_{t=0}^{T-1}\mathbb{E}\Big[{\big\lVert {\Omega}_K^{t}(Q^t - I_K) \big\rVert}^2 \Big] = 0.
\end{displaymath}
Assuming that $\mathop{lim}_{T \to \infty}{\big\lVert Q^T - I_K \big\rVert}^2 \neq 0$, we have
\begin{displaymath}
\exists j, k \in [K], \mathop{lim}_{T \to \infty}\big\lvert \big( Q^T - I_K \big)_{j,k} \big\rvert \neq 0.
\end{displaymath}
That is
\begin{displaymath}
\forall T \text{, } \exists j_T, k_T \in [K] \text{ and } {\delta}_T > 0 \text{, } \big\lvert \big( Q^T - I_K \big)_{j_T,k_T} \big\rvert > {\delta}_T.
\end{displaymath}
Because we can always find some ${\Omega}_K^{t}$ making that 
\begin{equation*}
\Big\lvert \sum_{j=1}^{K}({\Omega}_K^{t})_{l,j}(Q^t - I_K)_{j,k}\Big\rvert = \sum_{j=1}^{K}\big\lvert ({\Omega}_K^{t})_{l,j}(Q^t - I_K)_{j,k} \big\rvert,
\end{equation*}
we can get
\begin{equation*}
\begin{split}
&\quad \Big\lvert \sum_{t=0}^{T-1}{\big\lVert {\Omega}_K^{t}(Q^t - I_K) \big\rVert}^2 \Big\rvert
\\& = \sum_{t=0}^{T-1}\sum_{l=1}^{d}\sum_{k=1}^{K}{{\big[{\Omega}_K^{t}(Q^t - I_K)\big]}_{j,k}}^2
\\& = \sum_{t=0}^{T-1}\sum_{l=1}^{d}\sum_{k=1}^{K}{\Big[\sum_{j=1}^{K}({\Omega}_K^{t})_{l,j}(Q^t - I_K)_{j,k}\Big]}^2
\\& \geq \sum_{t=0}^{T-1}\sum_{l=1}^{d}\sum_{k=1}^{K}\Big[\sum_{j=1}^{K}{({\Omega}_K^{t})_{l,j}}^2{(Q^t - I_K)_{j,k}}^2\Big]
\\& \geq \sum_{t=0}^{T-1}\sum_{l=1}^{d}{({\Omega}_K^{t})_{l,j_t}}^2{(Q^t - I_K)_{j_t,k_t}}^2
\\& \geq \sum_{t=0}^{T-1}{{\delta}_{\Omega max}}^2{{\delta}_t}^2
\end{split}
\end{equation*}
where ${{\delta}_{\Omega max}}^2 = \mathop{min}_{t \in [T]}\mathop{max}_{l \in [d]}\big\{ {({\Omega}_K^{t})_{l,j_t}}^2\big\}$ and ${{\delta}_{\Omega max}}^2 > 0$ (Otherwise, $({\Omega}_K^{t})_{l,j_t} = 0, \forall l$. Thus, the $j_t$-th global model is invalid). Then we have
\begin{equation*}
\frac{1}{T}\sum_{t=0}^{T-1}\mathbb{E}\Big[{\big\lVert {\Omega}_K^{t}(Q^t - I_K) \big\rVert}^2 \Big] \geq \frac{1}{T}\sum_{t=0}^{T-1}{{\delta}_{\Omega max}}^2{{\delta}_t}^2 > 0.
\end{equation*}
That is
\begin{small}
\begin{equation*}
\forall T \text{, } \exists \delta = \frac{1}{T}\sum_{t=0}^{T-1}{{\delta}_{\Omega max}}^2{{\delta}_t}^2 >0 \text{, } \frac{1}{T}\sum_{t=0}^{T-1}\mathbb{E}\Big[{\big\lVert {\Omega}_K^{t}(Q^t - I_K) \big\rVert}^2 \Big] > \delta,
\end{equation*}
\end{small}which means that 
\begin{equation*}
\mathop{lim}_{T \to \infty}\frac{1}{T}\sum_{t=0}^{T-1}\mathbb{E}\Big[{\big\lVert {\Omega}_K^{t}(Q^t - I_K) \big\rVert}^2 \Big] \neq 0.
\end{equation*}
It contradicts the assumption. The proof of \textbf{" $\Longrightarrow$ "} ends.

\noindent \textbf{\itshape 1b)}\quad \textbf{" $\Longleftarrow$ ":} \quad We have 
\begin{equation*}
\mathop{lim}_{T \to \infty}{\big\lVert Q^T - I_K \big\rVert}^2 = 0,
\end{equation*}
which indicates that
\begin{small}
\begin{equation*}
\forall j, k \text{ and } {\varepsilon}_0 > 0 \text{, } \exists T_0 > 0 \text{, making } \forall T > T_0 \text{, } \big\lvert (Q^T - I_K)_{j,k} \big\rvert < {\varepsilon}_0. 
\end{equation*}
\end{small}We know that
\begin{equation*}
\mathop{lim}_{T \to \infty}\frac{T_1}{T} = 0 \text{, } \forall T_1,
\end{equation*}
which means that
\begin{equation*}
\forall {\varepsilon}_1 > 0 \text{, } \exists T_2 \text{, making } \forall T > T_2 \text{, } \frac{T_0 + 1}{T} < {\varepsilon}_1. 
\end{equation*}
When $T_3 = \mathop{max}\{T_0, T_2\}$, $\forall T > T_3$, we have
\begin{small}
\begin{equation*}
\begin{split}
&\quad \frac{1}{T}\sum_{t=0}^{T_0}{\big\lVert {\Omega}_K^{t}(Q^t - I_K) \big\rVert}^2
\\&= \frac{1}{T}\sum_{t=0}^{T_0}\sum_{l=1}^{d}\sum_{k=1}^{K}{\Big[\sum_{j=1}^{K}({\Omega}_K^{t})_{l,j}(Q^t - I_K)_{j,k}\Big]}^2
\\&= \frac{1}{T}\sum_{t=0}^{T_0}\sum_{l=1}^{d}\sum_{k=1}^{K} \Big[\sum_{j=1}^{K}{({\Omega}_K^{t})_{l,j}{(Q^t)}_{j,k}} - \sum_{j=1}^{K}{({\Omega}_K^{t})_{l,j}{(I_K)}_{j,k}} \Big]^2
\\&\leq \frac{2}{T}\sum_{t=0}^{T_0}\sum_{l=1}^{d}\sum_{k=1}^{K}\Big\{ \Big[\sum_{j=1}^{K}{({\Omega}_K^{t})_{l,j}(Q^t)_{j,k}}\Big]^2 
\\&\quad+ \Big[\sum_{j=1}^{K}{({\Omega}_K^{t})_{l,j}(I_K)_{j,k}}\Big]^2 \Big\}
\\&\leq \frac{2}{T}\sum_{t=0}^{T_0}\sum_{l=1}^{d}\sum_{k=1}^{K} \Big\{ \sum_{j=1}^{K}{({\Omega}_K^{t})_{l,j}}^2(Q^t)_{j,k} + {({\Omega}_K^{t})_{l,k}}^2 \Big\}
\\&= \frac{2}{T}\sum_{t=0}^{T_0}\sum_{l=1}^{d}\sum_{j=1}^{K}{({\Omega}_K^{t})_{l,j}}^2\sum_{k=1}^{K}(Q^t)_{j,k} + \frac{2}{T}\sum_{t=0}^{T-1}\sum_{l=1}^{d}\sum_{k=1}^{K}{({\Omega}_K^{t})_{l,k}}^2
\\&\leq \frac{4}{T}\sum_{t=0}^{T_0}\sum_{l=1}^{d}\sum_{k=1}^{K}{({\Omega}_K^{t})_{l,k}}^2
\\&\leq \frac{4{\rho}_{\Omega}^2(T_0+1)}{T}
\end{split}
\end{equation*}
\end{small}So, we can get
\begin{equation*}
\begin{split}
&\quad \Big\lvert \frac{1}{T}\sum_{t=0}^{T-1}{\big\lVert {\Omega}_K^{t}(Q^t - I_K) \big\rVert}^2 \Big\rvert
\\&= \frac{1}{T}\sum_{t=0}^{T_0}{\big\lVert {\Omega}_K^{t}(Q^t - I_K) \big\rVert}^2 + \frac{1}{T}\sum_{t=T_0+1}^{T-1}{\big\lVert {\Omega}_K^{t}(Q^t - I_K) \big\rVert}^2
\\&\leq \frac{4{\rho}_{\Omega}^2(T_0+1)}{T} 
\\&\quad+ \frac{1}{T}\sum_{t=T_0+1}^{T-1}\sum_{l=1}^{d}\sum_{k=1}^{K} K \Big[\sum_{j=1}^{K}{({\Omega}_K^{t})_{l,j}}^2{(Q^t - I_K)_{j,k}}^2\Big]
\\&\leq \frac{4{\rho}_{\Omega}^2(T_0+1)}{T} + \frac{1}{T}\sum_{t=T_0+1}^{T-1}\sum_{l=1}^{d}\sum_{k=1}^{K} K {{\varepsilon}_0}^2\sum_{j=1}^{K}{({\Omega}_K^{t})_{l,j}}^2
\\&\leq {\rho}_{\Omega}^2 \Big( \frac{4(T_0 + 1)}{T} + \frac{T - T_0 - 1}{T}K^2{{\varepsilon}_0}^2 \Big)
\\& < \underbrace{{\rho}_{\Omega}^2 \big(4{\varepsilon}_1 + K^2{{\varepsilon}_0}^2 \big)}_{:= \varepsilon}
\end{split}
\end{equation*}
That is
\begin{flalign*}
&\forall \varepsilon > 0 \text{, } \exists T_3 = \mathop{max}\{T_0, T_2\} \text{, making } \forall T > T_3 \text{, } &
\\&\qquad\qquad\qquad \Big\lvert \frac{1}{T}\sum_{t=0}^{T-1}{\big\lVert {\Omega}_K^{t}(Q^t - I_K) \big\rVert}^2 \Big\rvert < {\varepsilon}, &
\end{flalign*}
which is the definition of 
\begin{displaymath}
\mathop{lim}_{T \to \infty}\frac{1}{T}\sum_{t=0}^{T-1}\mathbb{E}\Big[{\big\lVert {\Omega}_K^{t}(Q^t - I_K) \big\rVert}^2 \Big] = 0.
\end{displaymath}
Thus, the proof of \textbf{" $\Longleftarrow$ "} ends.

From the analysis of the algorithm \textit{CGPFL}, we know the iterates of the global models are
\begin{equation*}
  {\Omega}_{K}^{0} \longrightarrow \cdots 
  \longrightarrow {\Omega}_K^t
  \longrightarrow {\Omega}_K^{t+1}
\end{equation*}
At any global round $t$, we consider a client $i$ which belongs to the cluster $k$ at current round, i.e., $i \in C_k^t$. At the next round $t+1$, we focus on any cluster $j$, where $j \in [K]$. According to the definition of $P^t$, we have
\begin{equation}
P^{t+1}_{i,j} = \sum_{p=1}^{K}P^t_{i,p}(Q^t)_{p,j}
\end{equation}
Since we focus on the disjoint cluster structure, i.e., $P_{i,k}^{t}=
\begin{cases}
\frac{1}{\lvert C_k^t \rvert}, & \mbox{if } i \in C_k^t \\
0, & \mbox{otherwise}
\end{cases}$, we can get that $P^{t+1}_{i,j} = \frac{1}{\lvert C_k^t \rvert}(Q^t)_{k,j}$. We know that the $k$-means clustering partitions the data points into different groups according to the distances between the data points and the centers of the clusters, i.e., $P^t_{i,k}=Probability\big(k=\mathop{\arg}\min_{p\in [K]}{\lVert \omega_{i,R}^{t-1}-\omega_p^{t} \rVert}^2\big)$.
% Using the iterates of global models, we can calculate the distance:
% \begin{equation*}
% \begin{split}
% &\quad {\big\lVert \omega_{i,R}^t - \omega_j^{t+1} \big\rVert}^2
% \\&= {\Big\lVert \omega_k^t-\beta\sum_{r=0}^{R-1}h_{i,r}^t - \Big[ \omega_j^t - \alpha(\omega_j^t-\frac{1}{\Big\lvert C_j^{t+1}\rvert}\sum_{p\in C_j^{t+1}}\beta\sum_{r=0}^{R-1}h_{p,r}^t) \Big] \Big\rVert}^2
% \\&= {\Big\lVert \omega_k^t-\omega_j^t + \alpha\Big( \omega_j^t-\frac{1}{\lvert C_j^{t+1} \rvert}\sum_{p\in C_j^{t+1}}\beta\sum_{r=0}^{R-1}h_{p,r}^t \Big) - \beta\sum_{r=0}^{R-1}h_{i,r}^t \Big\rVert}^2
% \\&= {\Big\lVert (1-\alpha)(\omega_k^t-\omega_j^t) + (\omega_{i,R}^t - \omega_k^{t+1}) + \alpha\beta\sum_{r=0}^{R-1}\Big[\frac{1}{\lvert C_j^{t+1}\rvert}\sum_{p\in C_j^{t+1}}h_{p,r}^t - \frac{1}{\lvert C_k^{t+1}\rvert}\sum_{p\in C_k^{t+1}}h_{p,r}^t\Big] \Big\rVert}^2
% \end{split}
% \end{equation*}
Because the global models are initialized from a same point, under the non-IID case, the distances between these models will necessarily become larger than certain tiny positive constants $\delta_d^2$ after one global steps. Then the models can be separated into different clusters, and gradually the cluster structure will remain invariant since the updates of model parameters become smaller and smaller as the learning rate shrinks. Therefore, as long as the index of the selected initialization centroid points in $k$-means clustering keeps unchange (e.g., $k$-means++. This is the reason why we adopt $k$-means++ in our algorithm to conduct clustering) during the algorithm, $Q^t$ will keep equal to $I_K$ after the first few global rounds. And we can get
\begin{equation}
\label{equ:B1}
  \frac{1}{T}\sum_{t=0}^{T-1}\mathbb{E}\Big[{\big\lVert {\Omega}_K^{t}(Q^t - I_K) \big\rVert}^2 \Big] \leq \mathcal{O}\Big(\frac{4{\rho}_{\Omega}^2}{T}\Big)
\end{equation}
Similarly, we can obtain
\begin{equation}
\label{equ:B2}
  \frac{1}{T}\sum_{t=0}^{T-1}\mathbb{E}\Big[{\big\lVert \nabla G_K({\Omega}_K^t)(Q^t - I_K) \big\rVert}^2 \Big] \leq \mathcal{O}\Big(\frac{4{\rho}_g^2}{T}\Big)
\end{equation}

In the next part, we will first deal with $\mathbf{B} = \mathbb{E}\Big[\sum_{k=1}^{K}\big[G_I({\Omega}_K^{t+1})P^t(Q^t - I_K) \big]_k \Big]$ and give the proof of $\mathbf{B} = 0$.
%\noindent{\itshape Proof:}
\begin{flalign*}
&\quad \sum_{k=1}^{K}\big[G_I({\Omega}_K^{t+1})P^t(Q^t - I_K) \big]_k &
\\&= \sum_{k=1}^{K}\sum_{j=1}^{K}\big[G_I({\Omega}_K^{t+1})P^t\big]_j(Q^t - I_K)_{j,k} &
\\&= \sum_{j=1}^{K}\big[G_I({\Omega}_K^{t+1})P^t\big]_j\sum_{k=1}^{K}(Q^t - I_K)_{j,k} &
\\&= \sum_{j=1}^{K}\big[G_I({\Omega}_K^{t+1})P^t\big]_j \Big[ \sum_{k=1}^{K}(Q^t)_{j,k} - \sum_{k=1}^{K}(I_K)_{j,k} \Big] \equiv 0, &
\end{flalign*}
no matter what value $G_I({\Omega}_K^{t+1})P^t$ takes. Therefore,
\begin{equation}
\mathbf{B} = \mathbb{E}\Big[\sum_{k=1}^{K}\big[G_I({\Omega}_K^{t+1})P^t(Q^t - I_K) \big]_k \Big] = 0.
\end{equation}
In conclusion,
\begin{flalign*}
  &\mathbb{E}\Big[\sum_{k=1}^{K}G_k({\omega}_k^{t+1}) - \sum_{k=1}^{K}G_k({\omega}_k^t) \Big] &
  \\&\leq -\frac{\hat{\alpha}}{4}\mathbb{E}\Big[{\big\lVert \nabla G_K({\Omega}_K^t)\big\rVert}^2\Big] + \frac{2{\alpha}^2}{\hat{\alpha}}\mathbf{B_1} + \frac{3\hat{\alpha}}{2}\mathbf{B_2} &
  \\&\quad+ \frac{208{\hat{\alpha}}^3{\delta}^2{L_G}^2}{NR{\alpha}^2} + \frac{13\hat{\alpha} {\delta}^2}{N}. &
\end{flalign*}
Reformulating it, we can get
\begin{flalign*}
  &\frac{1}{T}\sum_{t=0}^{T-1}\mathbb{E}\Big[\frac{1}{K}{\big\lVert \nabla G_K({\Omega}_K^t)\big\rVert}^2\Big] &
  \\&\leq \frac{4}{\hat{\alpha} T}\sum_{t=0}^{T-1}\mathbb{E}\Big[\frac{1}{K}\sum_{k=1}^{K}G_k({\omega}_k^{t}) - \frac{1}{K}\sum_{k=1}^{K}G_k({\omega}_k^{t+1}) \Big] &
  \\&\quad + \frac{8{\alpha}^2}{K{\hat{\alpha}}^2 T}\sum_{t=0}^{T-1}\mathbf{B_1} + \frac{6}{KT}\sum_{t=0}^{T-1}\mathbf{B_2} + \frac{832{\hat{\alpha}}^2 {\delta}^2{L_G}^2}{KNR{\alpha}^2} + \frac{52{\delta}^2}{KN} &
  \\&\leq \frac{4\mathbb{E}\Big[\frac{1}{K}\sum_{k=1}^{K}G_k({\omega}_k^{0}) - \frac{1}{K}\sum_{k=1}^{K}G_k({\omega}_k^{T}) \Big]}{\hat{\alpha} T} &
  \\&\quad+ \frac{32{\alpha}^2 \rho_{\Omega}^2}{K{\hat{\alpha}}^2 T} + \frac{24 {\rho}_g^2}{KT} + \frac{832{\hat{\alpha}}^2 {\delta}^2{L_G}^2}{KNR{\alpha}^2} + \frac{52{\delta}^2}{KN} &
\end{flalign*}
We define that ${\Delta}_G := \mathbb{E}\Big[\frac{1}{K}\sum_{k=1}^{K}G_k({\omega}_k^{0}) - \frac{1}{K}\sum_{k=1}^{K}G_k({\omega}_k^{T}) \Big]$ which is a constant with finite value, $C_1 := \frac{32{\rho}_{\Omega}^2}{K}$, $C_2 := \frac{24{\rho}_g^2}{K}$ and $C_3 := \frac{832 {\delta}^2{L_G}^2}{KNR}$, then we get
\begin{small}
\begin{equation}
  \frac{1}{T}\sum_{t=0}^{T-1}\mathbb{E}\Big[\frac{1}{K}{\big\lVert \nabla G_K({\Omega}_K^t)\big\rVert}^2\Big] \leq \frac{4{\Delta}_G}{\hat{\alpha} T} + \frac{C_1 {\alpha}^2}{{\hat{\alpha}}^2 T} + \frac{C_2}{T} + \frac{C_3{\hat{\alpha}}^2}{{\alpha}^2} + \frac{52{\delta}^2}{KN}.
\end{equation}
\end{small}With ${\hat{\alpha}}_0 := \mathop{min}\Big\{ \frac{C_1 {\alpha}^2}{4{\Delta}_G}, \sqrt{\frac{C_1}{C_2}}\alpha, \sqrt{\frac{1}{416{L_G}^2}}\alpha \Big\} $, we consider two cases as~\cite{karimireddy20SCAFFOLD,arjevani20converg1,dinh20pFedMe} do.

\noindent \textbf{If} ${\hat{\alpha}}_0 \leq \alpha\Big({\frac{C_1}{C_3 T}}\Big)^{\frac{1}{4}}$, we choose $\hat{\alpha} = {\hat{\alpha}}_0$. Thus we have
\begin{equation}
  \frac{1}{2T}\sum_{t=0}^{T-1}\mathbb{E}\Big[\frac{1}{K}{\big\lVert \nabla G_K({\Omega}_K^t)\big\rVert}^2\Big] \leq \frac{3 C_1 {\alpha}^2}{2{\hat{\alpha}_0}^2 T} + \frac{(C_1 C_3)^{\frac{1}{2}}}{2\sqrt{T}} + \frac{26{\delta}^2}{KN}.
\end{equation}

\noindent \textbf{If} ${\hat{\alpha}}_0 \geq \alpha\Big({\frac{C_1}{C_3 T}}\Big)^{\frac{1}{4}}$, we choose $\hat{\alpha} = \alpha\Big({\frac{C_1}{C_3 T}}\Big)^{\frac{1}{4}}$. Thus we have
\begin{equation}
\begin{split}
  \frac{1}{2T}\sum_{t=0}^{T-1}\mathbb{E}\Big[\frac{1}{K}{\big\lVert \nabla G_K({\Omega}_K^t)\big\rVert}^2\Big] &\leq \frac{3 C_1 {\alpha}^2}{2{\hat{\alpha}}^2 T} + \frac{C_3{\hat{\alpha}}^2}{2{\alpha}^2} + \frac{26{\delta}^2}{KN}
  \\&= \frac{2(C_1 C_3)^{\frac{1}{2}}}{\sqrt{T}} + \frac{26{\delta}^2}{KN}.
\end{split}
\end{equation}
Combining these two cases, we can obtain
\begin{equation}
\begin{split}
  &\frac{1}{T}\sum_{t=0}^{T-1}\mathbb{E}\Big[\frac{1}{K}{\big\lVert \nabla G_K({\Omega}_K^t)\big\rVert}^2\Big] \leq \frac{3 C_1 {\alpha}^2}{2{\hat{\alpha}_0}^2 T} + \frac{5(C_1 C_3)^{\frac{1}{2}}}{2\sqrt{T}} + \frac{52{\delta}^2}{KN}
  \\&\leq \frac{3 C_1 {\alpha}^2}{2{\hat{\alpha}_0}^2 T} + \frac{80\sqrt{26{\delta}^2{L_G}^2({\rho_{\Omega}^2}/K)}}{\sqrt{KNRT}} + \frac{52{\delta}^2}{KN}
\end{split}
\end{equation}

\noindent{\itshape Proof ends.}

As regard to the relationship between the personalized models and the global models, we adpot the process of the correspanding proof in~\cite{dinh20pFedMe}, and can get that
\begin{flalign*}
&\frac{1}{NT}\sum_{i=1}^{N}\sum_{t=0}^{T-1}\mathbb{E}\Big[ {\big\lVert \tilde{\theta}_i^t - \omega_{j}^t \big\rVert}^2\Big] &
\\&\leq \mathcal{O}\Big( \frac{1}{T}\sum_{t=0}^{T-1}\mathbb{E}\Big[\frac{1}{K}{\big\lVert \nabla G_K({\Omega}_K^t)\big\rVert}^2\Big]\Big) + \mathcal{O}\Big( \frac{\delta_G^2}{\lambda^2} + \delta^2 \Big) &
\end{flalign*}

\section{Proof of Generalization Bound}
Before we start the proof of the generalization bound, we first give some definitions which will be used in the following proof.
\begin{equation}
\begin{split}
  &h=h(\theta), \quad g=g(\omega)
  \\& \hat{h}_i^{\ast} = \hat{h}_i(\theta_i^{\ast}) = \mathop{\arg\min}\limits_{\theta_i}\big\{ \mathcal{L}_{\hat{D}_i}\big(h(\theta_i)\big) + \frac{\lambda}{2} {\lVert \theta_i - \omega_k^{\ast}\rVert}^2\big\}
  \\& h_i^{\ast} = h_i(\theta_i^{\ast}) = \mathop{\arg\min}\limits_{\theta_i}\big\{ \mathcal{L}_{D_i}\big(h(\theta_i)\big) + \frac{\lambda}{2} {\lVert \theta_i - \omega_k^{\ast}\rVert}^2\big\}
  \\& \hat{h}_{i,loc}^{\ast} = \hat{h}_{i,loc}(\theta_{i,loc}^{\ast}) = \mathop{\arg\min}\limits_{\theta_{i,loc}}\big\{ \mathcal{L}_{\hat{D}_i}\big(h(\theta_{i,loc})\big) \big\}
  \\& h_{i,loc}^{\ast} = h_{i,loc}(\theta_{i,loc}^{\ast}) = \mathop{\arg\min}\limits_{\theta_{i,loc}}\big\{ \mathcal{L}_{D_i}\big(h(\theta_{i,loc})\big) \big\}
\end{split}
\end{equation}
We can bound the generalization error of the obtained personalized models $\theta_i^{\ast} \text{, } i \in [N]$ by
\begin{small}
\begin{flalign*}
  &\quad \sum_{i=1}^{N}\frac{m_i}{m}\Big\{ \mathcal{L}_{D_i}(\hat{h}_i^{\ast}) - \min_{h\in \mathcal{H}}\mathcal{L}_{D_i}(h) \Big\} &
  \\&= \sum_{i=1}^{N}\frac{m_i}{m}\Big\{ \mathcal{L}_{D_i}(\hat{h}_i^{\ast}) - \mathcal{L}_{D_i}(h_{i,loc}^{\ast}) \Big\} &
  \\&= \sum_{i=1}^{N}\frac{m_i}{m}\Big\{ \mathcal{L}_{D_i}(\hat{h}_i^{\ast}) - \mathcal{L}_{D_i}(\hat{g}_k^{\ast}) + \mathcal{L}_{D_i}(\hat{g}_k^{\ast}) - \mathcal{L}_{\hat{D}_i}(\hat{g}_k^{\ast}) + \mathcal{L}_{\hat{D}_i}(\hat{g}_k^{\ast}) &
  \\&\quad- \mathcal{L}_{\hat{D}_i}(\hat{h}_i^{\ast}) + \mathcal{L}_{\hat{D}_i}(\hat{h}_i^{\ast}) - \mathcal{L}_{D_i}(\hat{h}_i^{\ast}) + \mathcal{L}_{D_i}(\hat{h}_i^{\ast}) - \mathcal{L}_{D_i}(h_{i,loc}^{\ast}) \Big\} &
  \\&= \sum_{i=1}^{N}\frac{m_i}{m}\Big\{ \mathcal{L}_{D_i}(\hat{g}_k^{\ast}) - \mathcal{L}_{\hat{D}_i}(\hat{g}_k^{\ast}) \Big\} + \sum_{i=1}^{N}\frac{m_i}{m}\Big\{ \mathcal{L}_{\hat{D}_i}(\hat{g}_k^{\ast}) - \mathcal{L}_{\hat{D}_i}(\hat{h}_i^{\ast}) \Big\} &
  \\&\quad+ \sum_{i=1}^{N}\frac{m_i}{m}\Big\{ \mathcal{L}_{D_i}(\hat{h}_i^{\ast}) - \mathcal{L}_{D_i}(\hat{g}_k^{\ast}) \Big\} &
  \\&\quad+ \sum_{i=1}^{N}\frac{m_i}{m}\Big\{ \mathcal{L}_{\hat{D}_i}(\hat{h}_i^{\ast}) - \mathcal{L}_{D_i}(h_{i,loc}^{\ast}) \Big\}
\end{flalign*}
\end{small}The above function is divided into four parts. In the following section, we will bound them sequentially.
To deal with the first part, we define that $k=\psi(i)$, where $i\in[N]$ and $k\in [K]$.
\begin{flalign*}
  &\sum_{i=1}^{N}\frac{m_i}{m}\Big\{ \mathcal{L}_{D_i}(\hat{g}_k^{\ast}) - \mathcal{L}_{\hat{D}_i}(\hat{g}_k^{\ast}) \Big\} &
  \\&\leq \mathop{\max}_{g_1,...,g_K}\sum_{i=1}^{N}\frac{m_i}{m}\mathop{\max}_{\psi(i)}\Big\{ \mathcal{L}_{D_i}(\hat{g}_{\psi(i)}^{\ast}) - \mathcal{L}_{\hat{D}_i}(\hat{g}_{\psi(i)}^{\ast}) \Big\} &
  \\&\leq \mathop{\max}_{\psi}\mathop{\max}_{g_1,...,g_K}\sum_{i=1}^{N}\frac{m_i}{m}\Big\{ \mathcal{L}_{D_i}(\hat{g}_{\psi(i)}^{\ast}) - \mathcal{L}_{\hat{D}_i}(\hat{g}_{\psi(i)}^{\ast}) \Big\} &
\end{flalign*}
Since the results of $k$-means++ depend on the selection of the first initialization centroid, the possible number of clustering results is $N$. By the McDiarmid's inequality, with probability at least $1-\delta$, we have
 \begin{flalign*}
  &\mathop{\max}_{g_1,...,g_K}\sum_{i=1}^{N}\frac{m_i}{m}\Big\{ \mathcal{L}_{D_i}(\hat{g}_{\psi(i)}^{\ast}) - \mathcal{L}_{\hat{D}_i}(\hat{g}_{\psi(i)}^{\ast}) \Big\} &
  \\&\leq \mathbb{E}\Big[ \mathop{\max}_{g_1,...,g_K}\sum_{i=1}^{N}\frac{m_i}{m}\Big( \mathcal{L}_{D_i}(\hat{g}_{\psi(i)}^{\ast}) - \mathcal{L}_{\hat{D}_i}(\hat{g}_{\psi(i)}^{\ast}) \Big) \Big] + 2\sqrt{\frac{\log{\frac{N}{\delta}}}{m}}
\end{flalign*}
Utilizing the results in~\cite{mansour20three}, we can get
 \begin{flalign*}
  &\mathbb{E}\Big[ \mathop{\max}_{g_1,...,g_K}\sum_{i=1}^{N}\frac{m_i}{m}\Big( \mathcal{L}_{D_i}(\hat{g}_{\psi(i)}^{\ast}) - \mathcal{L}_{\hat{D}_i}(\hat{g}_{\psi(i)}^{\ast}) \Big) \Big] &
  \\&\leq\frac{1}{m}\mathbb{E}\Big\{ \sum_{k=1}^{K}\mathop{\max}_{g_k} \Big[m_{C_k}\Big( \mathcal{L}_{D_{C_k}}(g_k) - \mathcal{L}_{\hat{D}_{C_k}}(g_k) \Big) \Big]\Big\} &
  \\&\leq \sum_{k=1}^{K}\frac{m_{C_k}}{m}\mathfrak{R}_{D_{C_k},m_{C_k}}(\mathcal{H}) \leq \sqrt{\frac{dK}{m}\log{\frac{em}{d}}}
\end{flalign*}
Therefore, we can get
\begin{equation}
\sum_{i=1}^{N}\frac{m_i}{m}\Big\{ \mathcal{L}_{D_i}(\hat{g}_k^{\ast}) - \mathcal{L}_{\hat{D}_i}(\hat{g}_k^{\ast}) \Big\} \leq 2\sqrt{\frac{\log{\frac{N}{\delta}}}{m}} + \sqrt{\frac{dK}{m}\log{\frac{em}{d}}}.
\end{equation}
When Assumption 1 is satisfied, we know that $\mathcal{L}_{\hat{D}_i}(h(\omega))$ is $L$-Lipschitz smooth. Thus, we have
\begin{equation}
\mathcal{L}_{\hat{D}_i}(\hat{g}_k^{\ast}) - \mathcal{L}_{\hat{D}_i}(\hat{h}_i^{\ast}) \leq \big\langle \nabla \mathcal{L}_{\hat{D}_i}(\hat{h}_i(\theta_i^{\ast})), \omega_k^{\ast} - \theta_i^{\ast} \big\rangle + \frac{L}{2}{\big\lVert \theta_i^{\ast} - \omega_k^{\ast}\big\rVert}^2
\end{equation}
Because $\hat{h}_i^{\ast}(\theta_i^{\ast})$ is obtained by solving $h_i(\theta_i^{\ast}) = \mathop{\arg\min}\limits_{\theta_i}\big\{ \mathcal{L}_{D_i}\big(h(\theta_i)\big) + \frac{\lambda}{2}{\lVert \theta_i - \omega_k^{\ast}\rVert}^2\big\}$, we can get that $\nabla \mathcal{L}_{\hat{D}_i}(\hat{h}_i(\theta_i^{\ast})) + \lambda(\theta_i^{\ast} - \omega_k^{\ast}) = 0$, that is $\nabla \mathcal{L}_{\hat{D}_i}(\hat{h}_i(\theta_i^{\ast})) = - \lambda(\theta_i^{\ast} - \omega_k^{\ast})$
Thus, we have
\begin{equation}
\sum_{i=1}^{N}\frac{m_i}{m}\Big\{ \mathcal{L}_{\hat{D}_i}(\hat{g}_k^{\ast}) - \mathcal{L}_{\hat{D}_i}(\hat{h}_i^{\ast}) \Big\} \leq (\lambda + \frac{L}{2})\sum_{i=1}^{N}\frac{m_i}{m} {\big\lVert \theta_i^{\ast} - \omega_k^{\ast} \big\rVert}^2
\end{equation}
Finally, according to the definitions of Complexity and Label-discrepancy, we can know that
\begin{small}
\begin{flalign*}
&\sum_{i=1}^{N}\frac{m_i}{m}\Big\{ \mathcal{L}_{D_i}(\hat{h}_i^{\ast}) - \mathcal{L}_{D_i}(\hat{g}_k^{\ast}) \Big\} + \sum_{i=1}^{N}\frac{m_i}{m}\Big\{ \mathcal{L}_{\hat{D}_i}(\hat{h}_i^{\ast}) - \mathcal{L}_{D_i}(h_{i,loc}^{\ast}) \Big\} &
\\&\leq 2B\sum_{i=1}^{N}\frac{m_i}{m}\mathop{\lambda_{\mathcal{H}}}(D_i) + \sum_{i=1}^{N}\frac{m_i}{m}\mathop{disc(D_i,\hat{D}_i)} &
\\&= \sum_{i=1}^{N}\frac{m_i}{m}\big\{ 2B\mathop{\lambda_{\mathcal{H}}}(D_i) + \mathop{disc(D_i,\hat{D}_i)} \big\}
\end{flalign*}
\end{small}where the constant $B$ satisfies that $\big\lvert\mathcal{L}_{D}(h_1) - \mathcal{L}_{D}(h_2)\big\rvert \leq B\mathop{\lambda_{\mathcal{H}}}(D)$ for $h_1 \text{, } h_2 \in \mathcal{H}$.
Summarizing the obtained results, we can get
\begin{flalign*}
&\sum_{i=1}^{N}\frac{m_i}{m}\Big\{ \mathcal{L}_{D_i}(\hat{h}_i^{\ast}) - \min_{h\in \mathcal{H}}\mathcal{L}_{D_i}(h) \Big\} &
\\&\leq 2\sqrt{\frac{\log{\frac{N}{\delta}}}{m}} + \sqrt{\frac{dK}{m}\log{\frac{em}{d}}} + (\lambda + \frac{L}{2})\sum_{i=1}^{N}\frac{m_i}{m} {\big\lVert \theta_i^{\ast} - \omega_k^{\ast} \big\rVert}^2 &
\\&\quad+ \sum_{i=1}^{N}\frac{m_i}{m}\big\{ 2B\mathop{\lambda_{\mathcal{H}}}(D_i) + \mathop{disc(D_i,\hat{D}_i)} \big\} &
\end{flalign*}
\section{More Experimental Details}
The dataset can be found via the following link: \\
\url{https://drive.google.com/file/d/1XqiMmJ9pI7apNfFlPwQFcW67rWvfD_aF/view?usp=sharing}.
\subsection{Convergence}

\begin{figure}[htbp]
  \centering
  \includegraphics[width=0.87\linewidth]{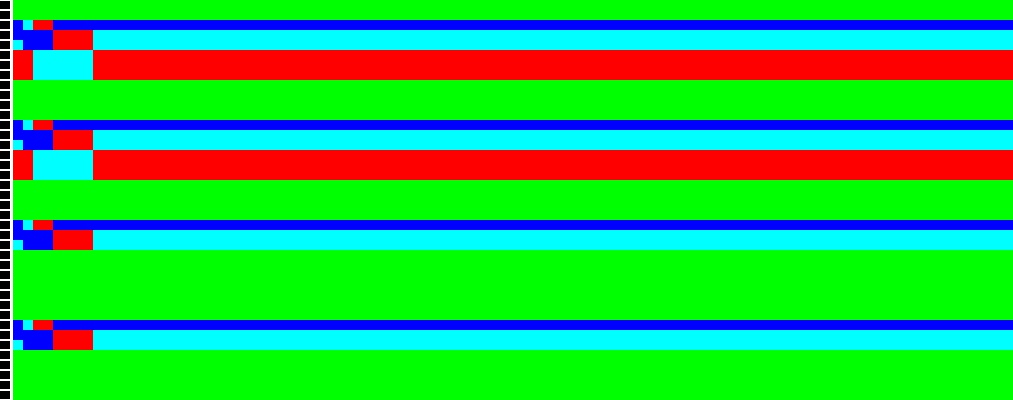}
  \caption{Convergence of the clustering results on model parameters.}
  \label{fig:belong_to}
  %\Description{}
\end{figure}

\begin{figure}[htbp]
  \centering
  \subfigure[The average distance among model parameters converges as the training proceeds.]{
    \begin{minipage}{0.22\textwidth}
    \centering
    \label{fig:models_dis}
    \includegraphics[width=1.1\textwidth]{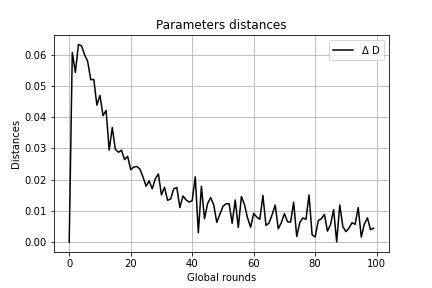}
    \end{minipage}
  }
  \subfigure[The overhead of K-means clustering at the server.]{
    \begin{minipage}{0.22\textwidth}
    \centering
    \label{fig:overhead}
    \includegraphics[width=1.1\textwidth]{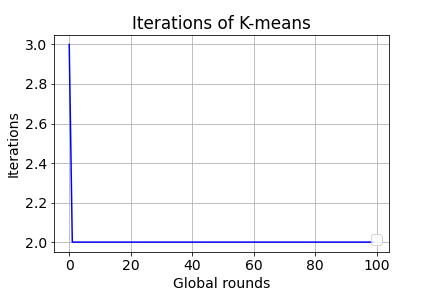}
    \end{minipage}
  }
  % \caption{Performance of \textit{CGPFL-Heur} on MNIST and Fashion-MNIST datasets}
  % \label{fig:cgpfl-heur}
  %\Description{}
\end{figure}

Finally, we provide some experimental results that support the convergence of the transition probability matrix $Q^t$ and show the overhead caused by {\itshape k}-Means clustering at the server.
Figure~\ref{fig:models_dis} demonstrates that the Euclidean distances between the models' parameters converge to a stable value as the training proceeds, which guarantees the convergence of the transition matrix $Q^t$ (the details can be found in the supplemental materials). 
Figure~\ref{fig:belong_to} shows the convergence of $Q^t$, where the horizontal axis indicate the iterations at the server, while each pixel in the vertical axis represents a client. The clients clustered into the same group at each iteration are painted the same color. We can see that the clustering result converges because the color map between clients gradually remains unchanged. 

% \begin{figure}[ht]
%   \centering
%   \includegraphics[width=0.9\linewidth]{figures/Models_distances.png}
%   \caption{The average distance among model parameters converges as the training proceeds.}
%   \label{fig:models_dis}
%   %\Description{}
% \end{figure}

% \begin{figure}[htbp]
%   \centering
%   \includegraphics[width=0.9\linewidth]{figures/km_iterations.png}
%   \caption{The overhead of K-means clustering at the server.}
%   \label{fig:clustering}
%   %\Description{}
% \end{figure}

The classic {\itshape k}-Means is a heuristic algorithm, of which the computation overhead is an unavoidable concern. In our method, on the one hand, the {\itshape k}-Means clustering is executed at the server which is usually considered having sufficient computing power. On the other hand, it can be observed from Figure~\ref{fig:overhead} that the the {\itshape k}-Means clustering can converge very fast with only few iterations after several global rounds. Therefore, the computation overhead caused by {\itshape k}-Means clustering is not a bottleneck in our method.

\subsection{The Effects of $\lambda$}
\begin{table}[htbp]
\centering
\scalebox{0.64}{
\begin{threeparttable}
\caption{Comparisons with various $\lambda$. We set $N=40$, $\alpha=1$, $R=10$, $S=5$, $lr=0.005$ and $T=200$ for MNIST and Fashion-MNIST (FMNIST), where $lr$ denotes the learning rate.}  
\label{tab:acc_lamda}
\setlength{\tabcolsep}{0.8mm}{    
\begin{tabular}{llccccccccc}  
\toprule
\multirow{5}{*}{MNIST-MLR}
% &$\lambda$ &$11$ &$12$ &$13$ &$14$ &$15$ &$16$ &$17$ &$18$ &$19$ &$20$ \cr
% &\textit{pFedMe (PM)} &$91.46$ &$91.90$ &$92.19$ &$92.54$ &$92.80$ &$93.00$ &$93.15$ &$93.16$ &$93.04$ &$92.66$ \cr
&$\lambda$ &$11$ &$12$ &$13$ &$14$ &$15$ &$16$ &$17$ &$18$ &$19$ \cr
&\textit{pFedMe (PM)} &$91.46$ &$91.90$ &$92.19$ &$92.54$ &$92.80$ &$93.00$ &$93.15$ &\underline{$93.16$} &$93.04$ \cr
&\textit{CGPFL (K=2)} &$93.43$ &$93.34$ &$93.62$ &$93.88$ &$94.16$ &$93.69$ &$93.52$ &$93.52$ &$93.31$ \cr
&\textit{CGPFL (K=4)} &$95.49$ &$95.65$ &$95.19$ &$95.47$ &$95.60$ &$95.77$ &$96.49$ &$94.85$ &$94.53$ \cr
&\textit{CGPFL-Heur}  &$97.46$ &$97.41$ &$96.27$ &$96.32$ &$96.34$ &$96.33$ &$96.32$ &$96.33$ &$96.25$ \cr
\midrule
\multirow{5}{*}{MNIST-DNN}
&$\lambda$ &$9$ &$10$ &$11$ &$12$ &$13$ &$14$ &$15$ &$16$ &$17$ \cr
&\textit{pFedMe (PM)} &$91.21$ &$91.54$ &$91.86$ &$92.21$ &$92.43$ &$92.79$ &$93.05$ &\underline{$93.30$} &$93.24$ \cr
&\textit{CGPFL (K=2)} &$94.11$ &$94.42$ &$94.71$ &$93.90$ &$94.14$ &$94.36$ &$94.49$ &$93.34$ &$93.36$ \cr
&\textit{CGPFL (K=4)} &$96.17$ &$96.37$ &$96.57$ &$96.55$ &$95.87$ &$95.99$ &$96.01$ &$95.45$ &$95.49$ \cr
&\textit{CGPFL-Heur}  &$97.69$ &$97.86$ &$98.00$ &$98.03$ &$97.95$ &$97.96$ &$98.20$ &$98.14$ &$98.16$ \cr
\midrule
\multirow{5}{*}{FMNIST-MLR}
&$\lambda$ &$9$ &$10$ &$11$ &$12$ &$13$ &$14$ &$15$ &$16$ &$17$ \cr
&\textit{pFedMe (PM)} &$85.03$ &$85.26$ &$85.42$ &$85.49$ &\underline{$85.49$} &$85.28$ &$85.16$ &$84.76$ &$84.22$ \cr
&\textit{CGPFL (K=2)} &$90.29$ &$87.70$ &$87.93$ &$88.00$ &$87.72$ &$87.53$ &$87.65$ &$86.94$ &$85.19$ \cr
&\textit{CGPFL (K=4)} &$92.50$ &$92.84$ &$92.94$ &$92.65$ &$92.63$ &$92.44$ &$92.42$ &$92.17$ &$92.20$ \cr
&\textit{CGPFL-Heur}  &$95.46$ &$95.44$ &$95.45$ &$95.36$ &$94.61$ &$94.40$ &$94.35$ &$94.41$ &$94.19$ \cr
\midrule
\multirow{5}{*}{FMNIST-DNN}
&$\lambda$ &$7$ &$8$ &$9$ &$10$ &$11$ &$12$ &$13$ &$14$ &$15$ \cr
&\textit{pFedMe (PM)} &$84.65$ &$85.20$ &$85.86$ &$86.28$ &$86.70$ &$86.87$ &$87.09$ &\underline{$87.10$} &$86.66$ \cr
&\textit{CGPFL (K=2)} &$87.69$ &$88.15$ &$88.72$ &$89.13$ &$89.59$ &$89.75$ &$91.15$ &$89.25$ &$88.93$ \cr
&\textit{CGPFL (K=4)} &$92.26$ &$92.89$ &$92.71$ &$92.86$ &$93.03$ &$93.56$ &$93.21$ &$93.44$ &$92.83$ \cr
&\textit{CGPFL-Heur}  &$95.60$ &$95.73$ &$95.84$ &$95.94$ &$95.98$ &$96.00$ &$95.98$ &$95.95$ &$95.83$ \cr

\bottomrule  
\end{tabular}
} 
%\footnotesize{The.}
\end{threeparttable}
}
%\footnotesize{The}
\vspace{-0.37cm}
\end{table}

As mentioned that the hyper-parameter $\lambda$ can balance the weight of personalization and generalization in several state-of-the-art PFL algorithms~\cite{dinh20pFedMe,hanzely20lower,li2021ditto}, we also conduct experiments to compare the performance of our \textit{CGPFL} and \textit{CGPFL-Heur} with a typical PFL algorithm, \textit{pFedMe}~\cite{dinh20pFedMe}, on different values of $\lambda$. 
Specifically, the range of $\lambda$ is properly chosen to avoid that divergence occurs in \textit{pFedMe}. 
%that divergence occurs in \textit{pFedMe}. 
The experimental results in Table~\ref{tab:acc_lamda} show that our methods can constantly achieve better performance than \textit{pFedMe} despite $\lambda$ varies, which demonstrates that \textit{CGPFL} can constantly reach better personalization-generalization trade-off against the state-of-the-art PFL methods.

\newpage

%% The file named.bst is a bibliography style file for BibTeX 0.99c
\bibliographystyle{named}
\bibliography{CGPFL-references}

\end{document}